%% file: main.tex
\PassOptionsToPackage{bookmarks=false}{hyperref} 
\documentclass[journal,twoside,web]{IEEEtran}
\usepackage[draft]{hyperref}
\usepackage{cite}
\usepackage{amsmath,amssymb,amsfonts}
\usepackage{graphicx}
\usepackage{textcomp}
\usepackage{subfigure}
\usepackage{lineno}
\usepackage{graphicx}
\usepackage{array}
\usepackage{multirow}
\usepackage{diagbox}
\usepackage{booktabs}
\usepackage{bm}
\usepackage{pifont}
\usepackage{gensymb}
\usepackage{algorithm}
\usepackage{algpseudocode}
\usepackage{amsmath}
\usepackage{graphicx}
\usepackage{subfigure}

\usepackage{siunitx}
\usepackage{xcolor}
\usepackage{bbding}
\usepackage{ulem}
\usepackage{mathrsfs}
\usepackage{lineno,hyperref}
\usepackage{indentfirst}
\usepackage{makecell}
\usepackage{color}
\usepackage{epstopdf}
\usepackage{textcomp}
\usepackage[export]{adjustbox}
\usepackage{url}

\newcommand{\ie}{\textit{i.e.,~}}
\newcommand{\etal}{\textit{et al.~}}
\newcommand{\eg}{\textit{e.g.,~}}

\newcommand{\specialcell}[2][l]{
    \begin{tabular}[#1]{@{}l@{}}#2\end{tabular}}

\newcommand{\revise}[1]{\textcolor{black}{#1}}
\newcommand{\wsrevise}[1]{\textcolor{black}{#1}}

\newcommand{\xre}[1]{\textcolor{black}{#1}}

\makeatletter
\newcommand\footnoteref[1]{\protected@xdef\@thefnmark{\ref{#1}}\@footnotemark}
\makeatother

\definecolor{darkgreen}{RGB}{50,150,50}
\begin{document}
\title{Learning Two-Stream CNN for Multi-Modal Age-related Macular Degeneration Categorization}
%\title{Learning Multi-Modal Stream CNNs for Age-related Macular Degeneration Categorization}
\author{Weisen Wang, Xirong Li, Zhiyan Xu, Weihong Yu, Jianchun Zhao, Dayong Ding, Youxin Chen
\thanks{This research was supported in part by NSFC (No. 62172420, 61672523), BJNSF (No. 4202033), BJNSF-Haidian Original Innovation Joint Fund (No. 19L2062), the Non-profit Central Research Institute Fund of Chinese Academy of Medical Sciences (No. 2018PT32029), CAMS Initiative for Innovative Medicine (CAMS-I2M, 2018-I2M-AI-001), and the Pharmaceutical Collaborative Innovation Research Project of Beijing Science and Technology Commission (No. Z191100007719002).  (\textit{Corresponding author: Xirong Li)}.}
\thanks{Weisen Wang  and Xirong Li are with the MoE Key Lab of Data Engineering and Knowledge Engineering, Renmin University of China and AIMC Lab, School of Information, Renmin University of China, Beijing 100872, China (e-mail: \{weisen,xirong\}@ruc.edu.cn)}
\thanks{Jianchun Zhao and Dayong Ding are with Vistel AI Lab, Visionary Intelligence Ltd., Beijing 100872, China  (e-mail: \{jianchun.zhao,dayong.ding\}@vistel.cn)}
\thanks{Zhiyan Xu, Weihong Yu and Youxin Chen are with the Key Lab of Ocular Fundus Disease, Chinese Academy of Medical Sciences and Dept. of Ophthalmology, Peking Union Medical College Hospital, Beijing 100730, China (e-mail: xuzythu@163.com, 536273640@qq.com, chenyouxinpumch@163.com)}
}

\markboth{IEEE Journal of Biomedical and Health Informatics,~Vol.~X, No.~X, XX~2022}%
{Wang \MakeLowercase{\textit{et al.}}: Learning Two-Stream CNN for Multi-Modal Age-related Macular Degeneration Categorization}

\maketitle

\begin{abstract}
This paper tackles automated categorization of Age-related Macular Degeneration (AMD), a common macular disease among people over 50. Previous research efforts mainly focus on AMD categorization with a single-modal input, let it be a color fundus photograph (CFP) or an OCT B-scan image. By contrast, we consider AMD categorization given a multi-modal input, a direction that is clinically meaningful yet mostly unexplored. Contrary to the prior art that takes a traditional approach of feature extraction plus classifier training that cannot be jointly optimized, we opt for end-to-end multi-modal Convolutional Neural Networks (MM-CNN). Our MM-CNN is instantiated by a two-stream CNN, with spatially-invariant fusion to combine information from the CFP and OCT streams. In order to visually interpret the contribution of the individual modalities to the final prediction, we extend the class activation mapping (CAM) technique to the multi-modal scenario.
For effective training of MM-CNN, we develop two data augmentation methods. One is GAN-based CFP/OCT image synthesis, with our novel use of CAMs as conditional input of a high-resolution image-to-image translation GAN. The other method is Loose Pairing, which pairs a CFP image and an OCT image on the basis of their classes instead of eye identities. Experiments on a clinical dataset consisting of 1,094 CFP images and 1,289 OCT images acquired from 1,093 distinct eyes show that the proposed solution obtains better F1 and Accuracy than multiple baselines for multi-modal AMD categorization. Code and data are available at \url{https://github.com/li-xirong/mmc-amd}.
\end{abstract}


\IEEEpeerreviewmaketitle

\section{Introduction}\label{sec:intro}
\input{intro}

\section{Related Work}\label{sec:related}
\input{relate}

\section{Proposed Solution}\label{sec:method}
\input{method}

\section{Experiments}\label{sec:eval}
\input{eval}

\section{Conclusions} \label{sec:conc}

\xre{T}his paper tackles multi-modal AMD categorization by end-to-end deep learning. Extensive experiments on a clinical dataset allow us to answer the questions asked in the beginning and consequently draw conclusions as follows. Compar\xre{ing} the two modalities, OCT is more effective than \xre{CFP}.
%for AMD categorization. 
When end-to-end trained, OCT-CNN obtains F1 of 0.886 and Accuracy of 0.818 on our dataset, making it a nontrivial baseline to beat. Compared to OCT-CNN, our multi-modal CNN  recognizes PCV and wetAMD with higher sensitivity and specificity, scoring F1 of 0.914 and Accuracy of 0.863. However, \xre{using} the two-stream network architecture alone is insufficient. In order to surpass the best single-modal baseline, the multi-modal CNN needs to be properly trained on data augmented by CAM-conditioned image synthesis and loose pairing. 

%\section*{Acknowledgments}

\appendix

\input{append}

\bibliographystyle{IEEEtran}
\bibliography{mybibfile}

\end{document}

%% file: intro.tex
%\IEEEPARstart{T}{his} paper targets at automated categorization of age-related macular degeneration (AMD). As a common macular disease among people over 50, AMD may cause blurred vision or even blindness if not treated in time \cite{Wan2014Global}. Depending on whether the retina contains choroidal neovascularization, AMD is classified into two subcategories, i.e., dry AMD (non-neovascular) and wet AMD (neovascular)~\cite{dewan2006htra1}. Moreover, when neovascularization occurs below the retinal pigment epithelium, such a type of wet AMD is known as Polypoidal Choroidal Vasculopathy (PCV)~\cite{lee2017efficacy,laude2010polypoidal,wong2016age-related}, typically characterized by polypoidal or aneurysmal dilations~\cite{kokame2019anti}. We refer to wet AMD excluding PCV as wetAMD, unless otherwise stated. Due to different treatments of dryAMD, PCV and wetAMD~\cite{wong2016age-related}, we need to not only distinguish AMD from normal, but also perform a fine-grained classification of the three AMD subcategories.  In the clinical practice, color fundus photography and optical coherence tomography (OCT) are widely used by an ophthalmologist to assess the condition of an eye. Examples of normal fundus and OCT images and those with specific AMD-related pathologies are shown in Fig. \ref{fig:fundus-oct-images}. Not surprisingly, the lack of experienced ophthalmologists has driven the research towards automated AMD categorization based on either color fundus images, OCT images or both. 

\begin{figure}[t!]
\centering
\includegraphics[width=\columnwidth]{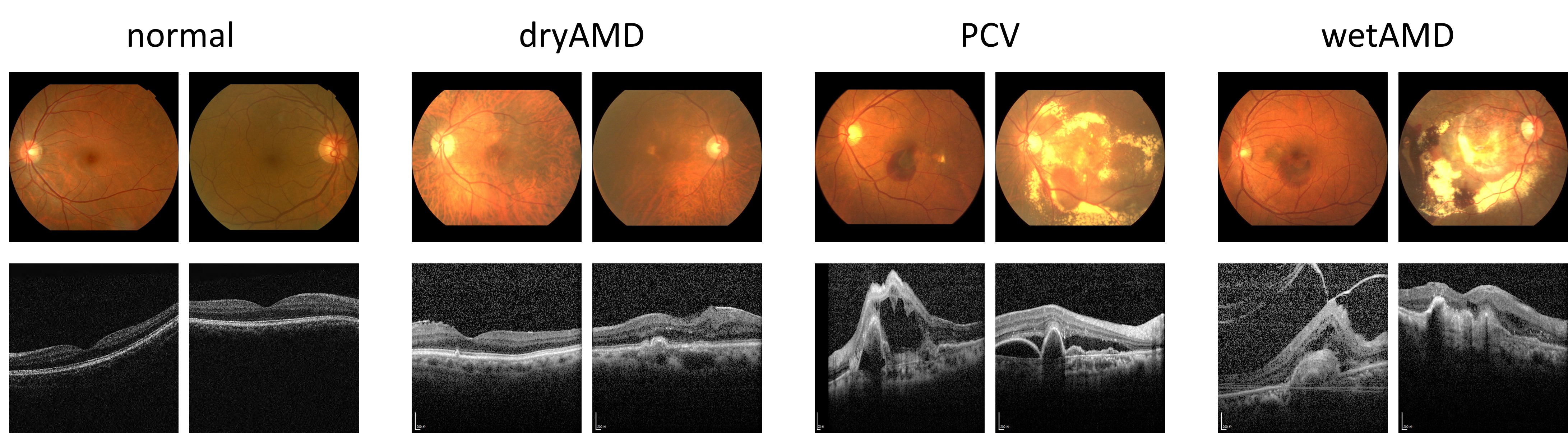}
\caption{\textbf{Color fundus images (first row) and OCT B-scan images (second row)} of normal eyes and eyes with specific diseases. Color fundus images show en face of the fundus, while OCT B-scans visualize cross-sectional information of the retina and choroid. Images per group are randomly selected from our dataset based on their classes.}
%rather than their eye identities}.}
\label{fig:fundus-oct-images}
\end{figure}

\IEEEPARstart{A}{ge}-related Macular Degeneration (AMD) is a common retinal disease among people over the age of 50. Without timely and proper treatment, patients with AMD may suffer from impaired vision or even blindness~\cite{Wan2014Global}. 
%Subject to the presence of choroidal neovascularization in the retina, the medical profession classifies the disease into \textit{dry} AMD (non-neovascular) and \textit{wet} AMD (neovascular)~\cite{dewan2006htra1}. 
Subject to the presence of choroidal neovascularization in the retina, the medical profession classifies the disease into \textit{dry} AMD (atrophic AMD and early stage AMD) and \textit{wet} AMD (neovascular AMD)~\cite{dewan2006htra1}.
Moreover, when neovascularization occurs below the retinal pigment epithelium, such a type of wet AMD is known as Polypoidal Choroidal Vasculopathy (PCV)~\cite{lee2017efficacy,laude2010polypoidal,wong2016age-related}, typically characterized by polypoidal or aneurysmal dilations~\cite{kokame2019anti}. For the ease of description, we refer to wet AMD excluding PCV as wetAMD hereafter. 
%, unless otherwise stated. 
Due to different treatments of dryAMD, PCV and wetAMD~\cite{wong2016age-related}, we need to not only distinguish AMD from normal, but also perform a fine-grained classification of the three AMD subcategories. 
To perform an AMD examination for a specific eye, an ophthalmologist commonly uses two noninvasive fundus imaging techniques, \ie color fundus photography (CFP) and optical coherence tomography (OCT). Given the retina, a CFP image shows its en face, while an OCT B-scan image captures its longitudinal section. 
%\ws{During the diagnosis, ophthalmologists often use color fundus photography and optical coherence tomography (OCT) to obverses the condition of eyes.} 
Examples of normal CFP and OCT images and those with specific AMD-related pathologies are shown in Fig. \ref{fig:fundus-oct-images}.
A recent study~\cite{de2018clinically} reports that when provided with multi-modal information (CFP / OCT images and clinical notes), experts make fewer diagnostic errors for referral recommendation. 
Unsurprisingly, due to the shortage of experienced ophthalmologists, there is a growing amount of research efforts towards automated AMD screening  based on either color fundus images~\cite{Burlina2016Detection,Burlina2017Automated,Grassmann2018A}, OCT images~\cite{Lee2017Deep,Karri2017Transfer,Treder2018Automated} or both~\cite{YooThe}.

%The majority of previous works are based on a single modality, let it be color fundus images capturing the posterior pole~\cite{kanagasingam2014progress,Burlina2016Detection,Burlina2017Comparing,Burlina2017Automated,Grassmann2018A} or OCT images of the macular~\cite{Lee2017Deep,Karri2017Transfer,Treder2018Automated,Russakoff2019Deep,Kermany2018Identifying}. In \cite{Burlina2016Detection}, for instance,  Burlina \etal employ a deep convolutional neural network (CNN) pretrained on ImageNet~\cite{deng2009imagenet} to extract visual features from fundus images and then train a linear SVM classifier. As for OCT-based methods, Lee \etal \cite{Lee2017Deep} train a VGG16~\cite{simonyan2014very} model to classify OCT images either as normal or as AMD. 
	
Existing works on AMD categorization mainly exploit one modality, let it be CFP images of the posterior pole~\cite{Burlina2016Detection,Burlina2017Comparing,Burlina2017Automated,Grassmann2018A} or OCT images of the macular~\cite{Lee2017Deep,Karri2017Transfer,Treder2018Automated,Russakoff2019Deep,Kermany2018Identifying}. In \cite{Burlina2017Automated}, for example, Burlina \etal deal with CFP, training AlexNet~\cite{krizhevsky2012imagenet}, a deep convolutional neural network (CNN), from scratch. An OCT-based method is developed by Lee \etal~\cite{Lee2017Deep}, where a VGGNet~\cite{simonyan2014very} is trained to predict the presence of AMD in OCT images.

%As for OCT-based research, Treder \cite{Treder2018Automated} fine-tune a deep convolutional neural network (CNN) pre-trained on ImageNet to classify the condition of an eye.}

%Since fundus images capture the state of the retinal plane, while OCT images reflect the longitudinal section of the retina, they describe distinct aspects of the retina, and  thus complementary to each other. While jointly exploiting the two modalities seems to be natural, this direction is largely unexplored. To the best of our knowledge, Yoo \etal \cite{YooThe} make an initial attempt towards multi-modal AMD categorization. 

%Given that color fundus and OCT images reflect the state of the retina in distinct aspects, and thus complement each other, a joint utilization of both modalities appears to be natural. To our surprise, however, few attempt is made in this direction. 
Given that CFP and OCT images reflect the state of the retina in distinct aspects, and thus complement each other, a joint utilization of both modalities appears to be natural. To our surprise, however, few attempt is made in this direction. 
%\ws{Color fundus images describe the retinal plane, whereas OCT images capture the longitudinal section of the retinal. These two modalities complementary to each other as they show different aspects of the retinal. Although it seems to be nature to jointly develop both of them, the direction is little explored.} 
To the best of our knowledge, the first work on multi-modal AMD categorization is by Yoo \etal \cite{YooThe}.  
%In a similar vein to \cite{Burlina2016Detection}, 
The authors leverage a pre-trained VGGNet to extract features from CFP and OCT images. The features are concatenated and classified by a random forest classifier. Their preliminary experiment indicates that the multi-modal method outperforms  its single-modal counterpart.
Nevertheless, given the rapid progress of deep learning for visual recognition, the VGGNet feature and the random forest classifier are both suboptimal.  \xre{We thus see multiple} crucial questions \xre{left} unanswered.
%Despite their encouraging result that the multi-modal method is better than its single-modal counterpart, we argue that some crucial questions remain open.
\xre{First}, 
 %Therefore, the first question arises as, 
 suppose the single-modal baseline is re-implemented based on a state-of-the-art CNN trained end-to-end for the task, will the current multi-modal method~\cite{YooThe} remain more effective?  If the answer turns out to be ``no'', further questions naturally follow as what deep learning network is suited for the task and how to train it effectively? Training a multi-modal network is nontrivial, as paired multi-modal training instances are far fewer than single-modal training instances. 
%\ws{if we reproduce the single-modal baseline based on a state-of-the-art CNN, such as ResNet \cite{He2016Deep}, and train it end-to-end, will the multi-modal proposal of Yoo \etal still be better? another question arises as, how to well implement an end-to-end multi-modal network for AMD categorization. It is not nontrivial to learn a multi-modal CNN, because the multi-modal input instances are image pairs which are less than the single-modal instances.} 
Therefore, one cannot take for granted that a multi-modal CNN will be better than its single-modal competitor. \xre{Last but not least, how to} %, 
%the method of~\cite{YooThe} lacks the ability of 
visually interpret the contribution of each modality to the final prediction? 
%\ws{Moreover, the proposal of Yoo \etal lacks the interpreting ability to view contribution of each modality to the final prediction.}

%Note that both the VGG19 features and the random forest classifier used in \cite{YooThe} are suboptimal in the context of deep learning based visual categorization. So the first question arises as, when the single-modal baseline is re-implemented using a state-of-the-art CNN, say ResNet \cite{He2016Deep}, and learned in an end-to-end manner, will the multi-modal method by \cite{YooThe} still be better? If the answer is negative, a follow-up question is can multi-modal AMD categorization be performed end-to-end as well? Training a deep learning network with multi-modal input is nontrivial, because by definition, the number of paired multi-modal training instances is less than the number of single-modal training instances. Therefore, one cannot take for granted that a multi-modal CNN will be better than its single-modal competitor. Last but not least, the method by \cite{YooThe} lacks the capability of interpreting how the individual modalities contribute to the final prediction.

\begin{figure*}[tbh!]
\begin{minipage}[b]{1.25\columnwidth}
    \subfigure[Two-stream CNN]{\label{fig:pipeline:a}
    \includegraphics[width=1\columnwidth]{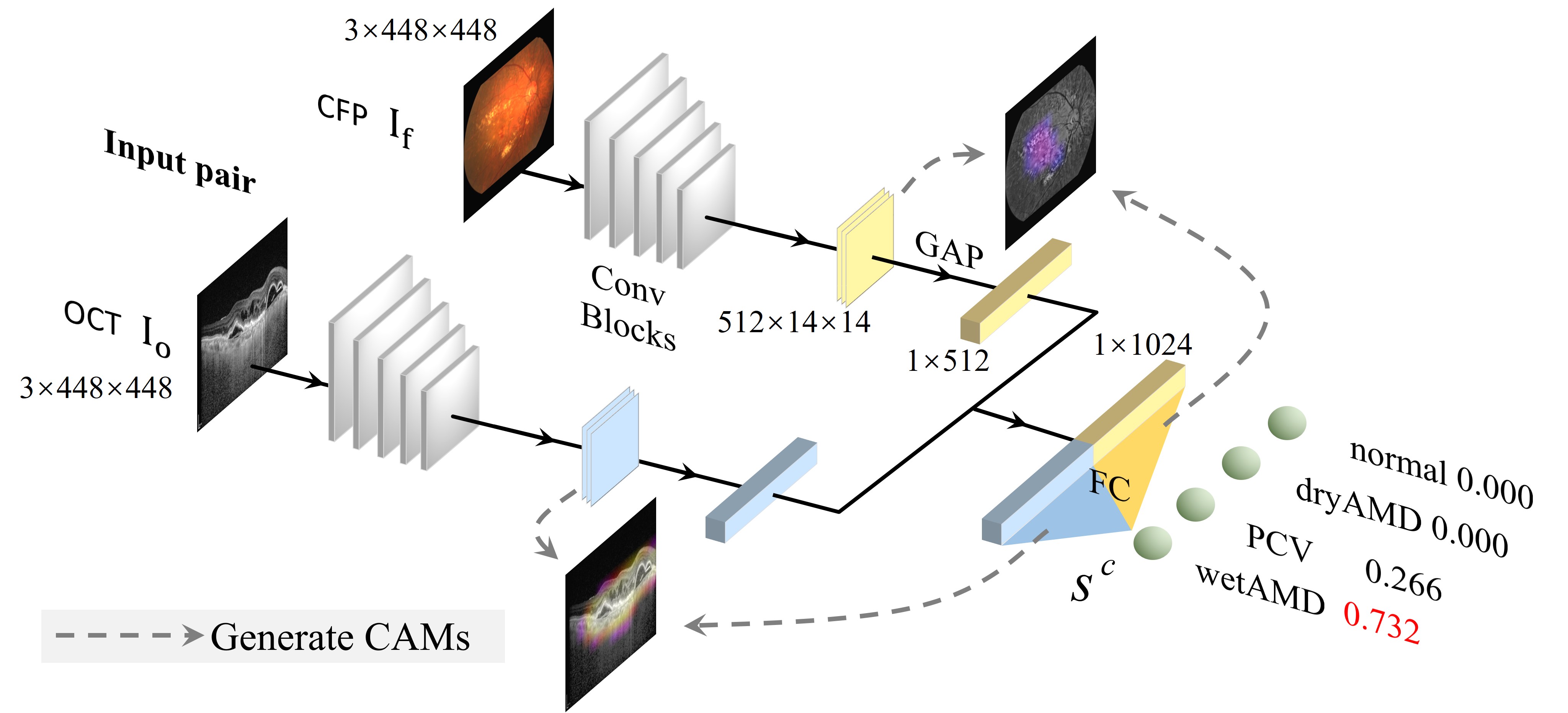}}
    \subfigure[CAM-conditioned image synthesis]{\label{fig:pipeline:b}
    \includegraphics[width=1\columnwidth]{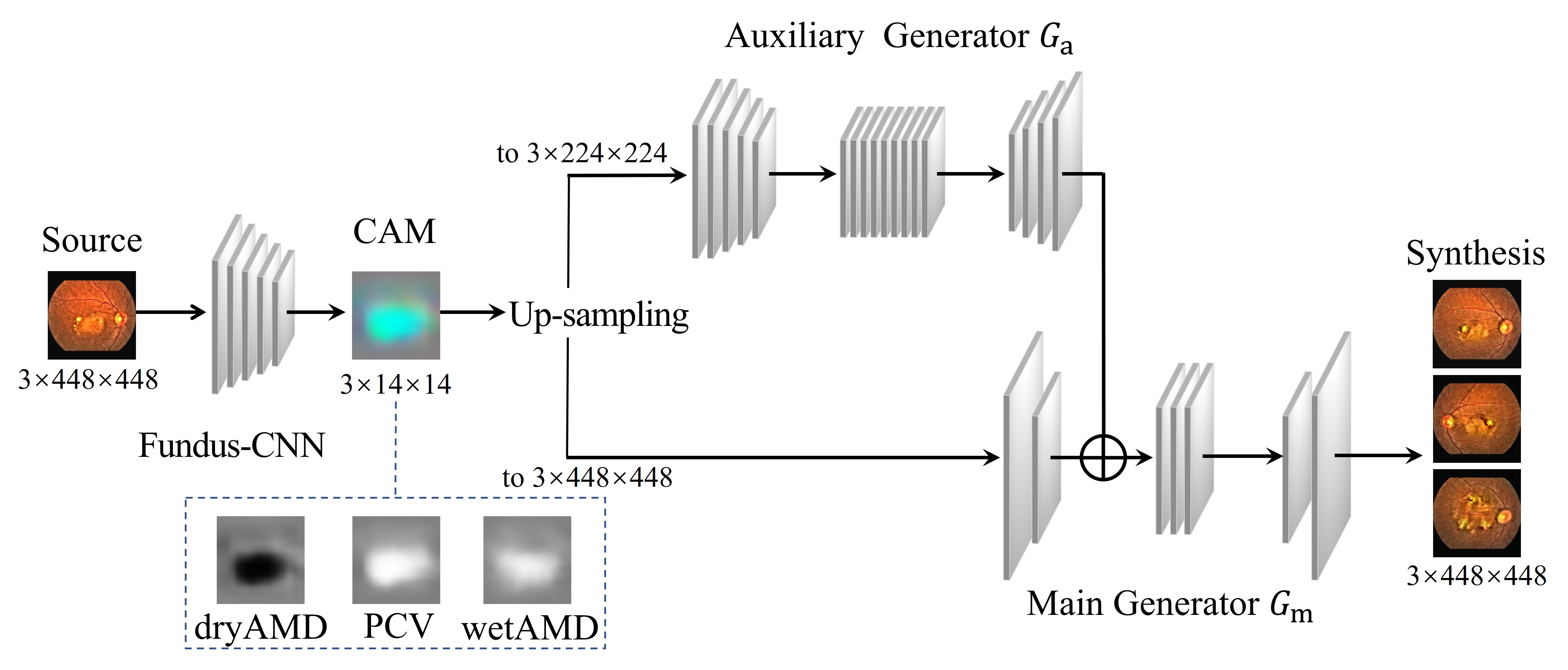}}
\end{minipage} %\par
\medskip
\begin{minipage}[b]{0.75\columnwidth}
    \subfigure[\xre{Two-stage} training pipeline]{\label{fig:pipeline:c}
    \includegraphics[width=0.9\columnwidth,right]{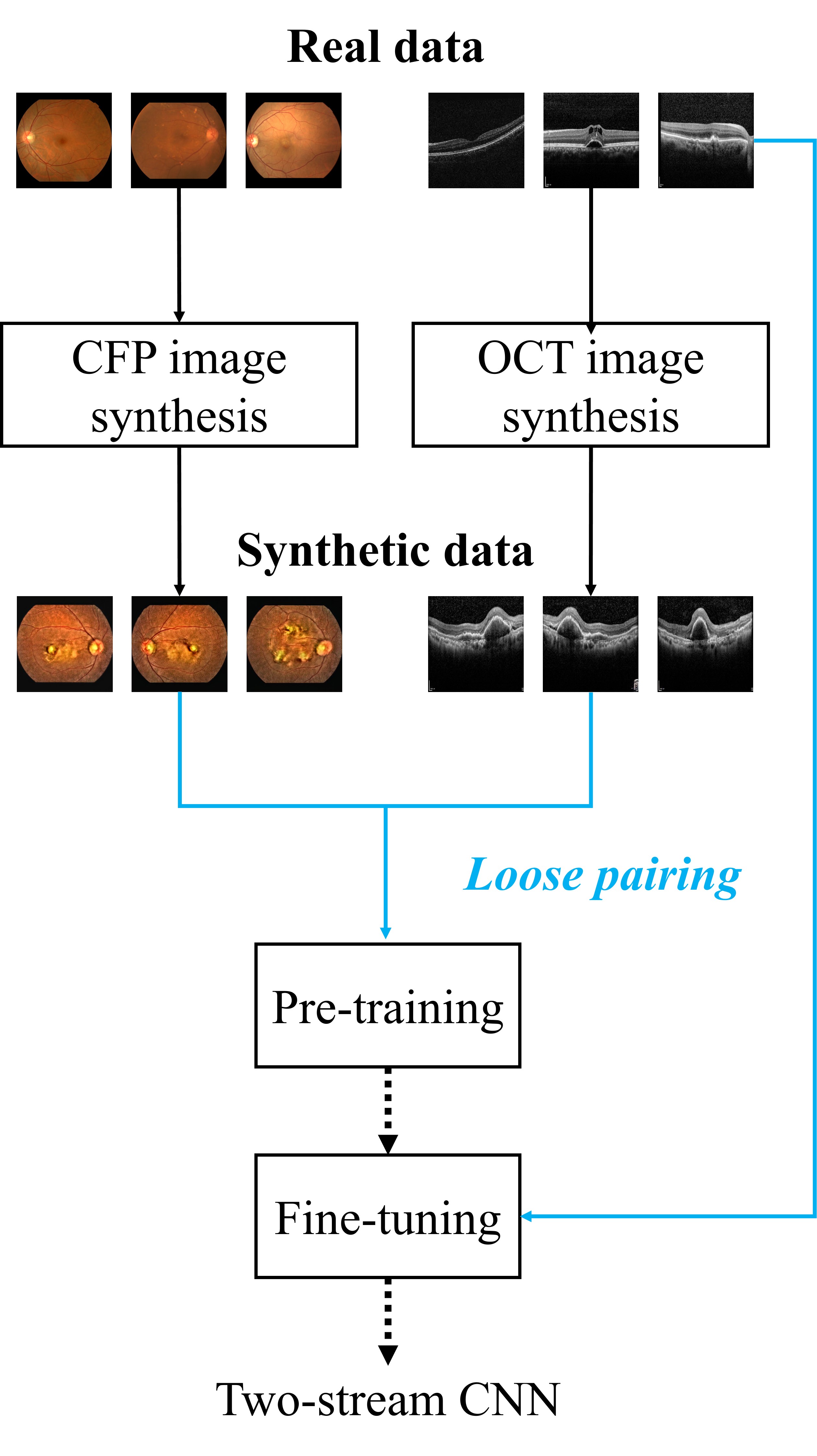}}
\end{minipage}
\caption{\textbf{Proposed end-to-end solution for multi-modal AMD categorization}. Given a pair of CFP and OCT images from a specific eye, our two-stream CNN makes a four-class prediction concerning the probability of the eye being \textit{normal}, \textit{dryAMD}, \textit{PCV} and \textit{wetAMD}, respectively. Extending class activation mapping (CAM)  to the multi-modal scenario allows us to visualize contributions of \xre{each} modality to final predictions. For effective training, we introduce two data augmentation methods. The first method is to synthesize CFP / OCT images by pix2pixHD, a high-resolution image-to-image translation network re-purposed in the new context. Given a  $3 \times 448 \times 448$ source image, we pre-train the corresponding single-modal CNN to produce CAMs \textit{w.r.t} each AMD class. The CAMs are stacked to form an three-channel image $[CAM^{dry}; CAM^{pcv}; CAM^{wet}]$ of $3 \times 14 \times 14$, which is then fed into pix2pixHD for image synthesis. A fully convolutional network $G_a$ is used as an auxiliary generator to generate a $3 \times 224 \times 224$ image. With the help of $G_a$, another fully convolutional network $G_m$ is then used as a main generator to generate a double-sized image. Manipulating the CAMs results in multiple synthesized images. The second method pairs CFP and OCT images based their classes instead of eye identities. A two-stage training is performed, where the two-stream CNN is first pre-trained on synthetic and loosely paired CFP and OCT images, and then fine-tuned on real and loosely paired multi-modal data.}
\label{fig:pipeline}
\end{figure*}

Towards answering the above questions, we make contributions as follows: \\
%\begin{itemize}
%\item 
$\bullet$ We propose a novel end-to-end solution for multi-modal AMD categorization, the first of its kind to the best of our knowledge. The backbone of our solution is a two-stream CNN, see Fig. \ref{fig:pipeline:a}. While such an architecture has been applied for video action recognition~\cite{feichtenhofer2016convolutional}, the fusion layer needs to be re-considered for the new task, not only for effectively fusing information from the individual modalities, but also for visualizing their contributions to the final prediction. To that end, we propose to use spatially-invariant fusion (SI-Fusion), which simply performs global average pooling on modal-specific feature maps followed by vector concatenation. SI-Fusion is not new by itself. Its novelty is due to its suitability for multi-modal AMD categorization. Compared to EarlyFusion which  concatenates CFP and OCT together as a whole, LateFusion which averages predictions of CFP- / OCT- CNNs, and multi-head self-attention based fusion \cite{yu2019multimodal}, SI-Fusion strikes a  proper balance between a network's learning capacity (for exploiting the complementarity between CFP and OCT) and its training complexity (for being more data-efficient). SI-Fusion also allows us to extend the CAM visualization technique \cite{Zhou2015Learning} for the first time to the multi-modal scenario for visual interpretation. \\
%SI-Fusion is thus a key component in our novel working solution}. 
%\item \ws{Targeting at multi-modal AMD categorization, we build a two-stream CNN in which different modalities are separately processed and then fused together for prediction, as shown in Fig. \ref{fig:pipeline}. The two-stream architecture is widely used in video action recognition \cite{feichtenhofer2016convolutional}. In this work, we re-consider the location of the fusion layer for not only effectively fusing the features from different modalities but also visualizing their contributions to predictions.}
%
%\item To attack the inadequacy of multi-modal paired instances for training, 
%\item 
$\bullet$ Due to the natural shortage of multi-modal training data, learning two-stream CNN with conventional data augmentation methods is insufficient. For multi-modal training data augmentation, we introduce two methods, \ie \textsl{CAM-conditioned Image Synthesis} and \textit{Loose Pairing}.  The first method, working at the image level, is to synthesize CFP / OCT images by a high-resolution generative adversarial network (GAN)~\cite{wang2018high} re-purposed in the new context, see Fig. \ref{fig:pipeline:b}. The second method pairs CFP and OCT images w.r.t their classes instead of eye identities. Consequently, we develop a training pipeline, as illustrated in Fig. \ref{fig:pipeline:c}, to effectively learn two-stream CNN from the augmented data. \\
%
%\item
$\bullet$ Experiments on a clinical dataset, with 1,094 CFP and 1,289 OCT images acquired from 1,093 eyes, verify the effectiveness of our solution. It surpasses the prior art~\cite{YooThe} clearly, with 0.914 \textit{vs} 0.792 in F1-score and 0.863 \textit{vs} 0.690 in Accuracy, for four-class AMD categorization. Code, data and model links are available at github\footnote{\url{https://github.com/li-xirong/mmc-amd}\label{github}}.

To sum up,  the novelty of this paper is two-fold, \ie clinically, the end-to-end multi-modal approach to AMD detection and technically, the CAM-based data augmentation with loose pairing. A preliminary version of this work was published at MICCAI'19 \cite{wang2019two}. We improve over the conference version in three aspects, \ie problem setup, method and evaluation. First, PCV is considered, making our setup more \xre{clinically meaningful}, yet more challenging due to its high resemblance to other wet AMD. Second, we enhance our data augmentation process with GAN-based CFP / OCT image synthesis (Sec. \ref{sssec:gan}). For an effective use of both real and synthetic training data, a new training pipeline is developed (Sec. \ref{ssec:training}). The above changes in problem setup and method naturally lead to a substantial extension of our evaluation. 

The rest of the paper is organized as follows. We discuss related work in Sec. \ref{sec:related}, followed by the proposed solution in Sec. \ref{sec:method},  experiments in Sec. \ref{sec:eval}, and conclusions in Sec. \ref{sec:conc}.

%% file: relate.tex
As this paper contributes to automated AMD categorization in terms of network designs and data augmentation methods, we review prior art for AMD categorization in Sec. \ref{ssec:related-amd} and data augmentation in Sec. \ref{ssec:related-da}. Note that we focus on deep learning approaches. For tackling AMD categorization by non-deep learning approaches, we refer to \cite{kanagasingam2014progress}.

\subsection{Deep Learning based AMD categorization} \label{ssec:related-amd}

\input{table_related}

%As we summarize in Table \ref{tab:related}, the majority of 
Existing works  are \xre{largely} based on a single modality, either \revise{CFP} images \cite{Burlina2016Detection,Burlina2017Comparing,Burlina2017Automated,Grassmann2018A} or OCT images \cite{Lee2017Deep,Karri2017Transfer,Treder2018Automated,Russakoff2019Deep}, \xre{see Table \ref{tab:related}}. In \cite{Burlina2016Detection,Burlina2017Comparing}, for instance, Burlina \etal tackle AMD categorization via a classical image classification approach, \ie feature extraction plus classifier training. In particular, the authors employ a CNN pretrained on ImageNet to extract visual features from \revise{CFP} images and then train a linear SVM classifier. In a follow-up work~\cite{Burlina2017Automated}, Burlina \etal find that an end-to-end trained CNN performs better than the classical approach. Ensemble learning is exploited by Grassmann \etal to combine multiple CNNs~\cite{Grassmann2018A}. As for OCT-based methods,  Lee \etal \cite{Lee2017Deep} train a VGG16 model from scratch to classify OCT images either as \textit{normal} or as \textit{AMD}. By contrast, Karri \etal \cite{Karri2017Transfer} and Treder \etal \cite{Treder2018Automated} train their CNNs by fine tuning. Different from the above works, we exploit both \revise{CFP} and OCT images as multi-modal input. 

Few attempt has been made for multi-modal AMD categorization. To the best of our knowledge, Yoo \etal \cite{YooThe} make the first effort in this direction, where the authors  perform three-class categorization, \ie normal, dryAMD and wetAMD, given \revise{CFP} and OCT images. Similar to \cite{Burlina2016Detection,Burlina2017Comparing} in the single-modal context, the authors follow the classical approach, employing a pre-trained VGG19 model to extract visual features from both modalities. The features are then concatenated and used as input of a random forest classifier. Different from \cite{YooThe}, our 
%two-stream CNN 
\xre{model is}
%can be 
trained end-to-end.
%in an end-to-end fashion. 

\subsection{Data Augmentation for CNN Training} \label{ssec:related-da}

Ever since the great success of AlexNet~\cite{krizhevsky2012imagenet}, data augmentation using low-level image processing techniques such as crop, flip, rotation and changes in brightness, saturation and contrast has been a rule of thumb for CNN training. Not surprisingly, we observe from Table \ref{tab:related} that data augmentation used by existing works for AMD categorization all follows this convention.  We notice a very recent attempt by Burlina \etal ~\cite{burlina2019assessment} to train a binary AMD classifier using GAN-generated examples. In particular, the authors train two PGGANs~\cite{karras2017progressive}, one for generating normal \revise{CFP} images and the other for synthesizing \revise{CFP} images with AMD. Their motivation, \ie to substitute GAN-generated data for manually labeled data, differs from ours. Moreover, according to their study, the classifier trained on synthetic data has lower performance than its counterpart trained on real data (0.8292 \textit{vs} 0.9112 in Accuracy and 0.9235 \textit{vs} 0.9706 in AUC). The result suggests that synthetic data alone is insufficient. 

\input{fig-gan}

For synthesizing high-resolution training images, there is an increasing interest in leveraging image-to-image GANs~\cite{zheng2018detection,shin2018medical,xing2019adversarial,zhou2019high,yi2019generative}. For detection of exudates in \revise{CFP} images, Zheng \etal \cite{zheng2018detection} expand their training data by developing a pix2pix~\cite{isola2017image} based network that converts exudate segmentation maps into \revise{CFP} images. As illustrated in Fig. \ref{fig:pix2pix}, a pix2pix model converts an input image, \eg an exudate segmentation map in \cite{zheng2018detection}, to a synthetic \revise{CFP} image by a fully convolutional network known as a generator. The generator is trained to fool another convolutional network known as a discriminator, which is responsible for discriminating between real and fake samples.  For tumor segmentation in brain MRI images, Shin \etal \cite{shin2018medical} train pix2pix to synthesize MRI images from tumor and tissue segmentation maps. For lesion detection in chest X-ray images, Xing \etal \cite{xing2019adversarial} also adopt pix2pix, where the input of pix2pix is obtained by masking out all regions of abnormal images except for lesion areas. In the context of diabetic retinopathy grading, Zhou \etal~\cite{zhou2019high} synthesize \revise{CFP} images by a pix2pixHD~\cite{wang2018high} based network, which translates vessel and lesion masks into \revise{the} images. As shown in Fig. \ref{fig:pix2pixhd}, the major improvement of pix2pixHD over pix2pix is a joint use of two generators and two discriminators to generate higher-resolution images in a coarse-to-fine manner.

%\ws{Since pix2pix and its high resolution variant pix2pixHD have shown effectiveness in the above studies, our proposed method uses pix2pixHD as well. Fig. \ref{fig:p2pgan} illustrates the architecture of the two models.} 

Our proposed method uses pix2pixHD. However, in contrast to existing works where the input is manually annotated~\cite{zheng2018detection,xing2019adversarial}, or automatically produced yet requiring manual annotation for training underlying semantic segmentation models~\cite{zhou2019high}, or both~\cite{shin2018medical}, our input, namely Class Activation Maps (CAMs), is automated extracted by image classifiers, with no need of any lesion annotation.

%% file: table_related.tex
\begin{table*}[tb!]
\renewcommand\arraystretch{1}
\centering
\caption{\textbf{State-of-the-art for automated AMD categorization}. This paper proposes two-stream CNN for multi-modal AMD categorization, and develops multi-modal data augmentation for effective training.}
\label{tab:related}
\scalebox{0.93}{
\begin{tabular}{@{}|l | l | l | l | @{}}
 \hline
\textbf{Modality}  & \textbf{Paper}  & \textbf{Categorization model} & \textbf{Data augmentation} \\
 \hline
\multirow{3}{*}{\textit{\xre{CFP}}}    & Burlina \etal \cite{Burlina2016Detection,Burlina2017Comparing} & \specialcell{OverFeat feature +  linear SVM classifier} & \specialcell{unmentioned} \\
 \cline{2-4}
  & Burlina \etal \cite{Burlina2017Automated} & \specialcell{AlexNet trained from scratch} & \specialcell{unmentioned}  \\
  \cline{2-4}
  & Grassmann \etal \cite{Grassmann2018A}     & \specialcell{Ensemble of AlexNet, GoogLeNet, Inception-V3, \\VGG11, ResNet-101, and Inception-ResNet-V2} & \specialcell{$\circ$ crop, flip, rotation} \\                    
 \hline
\multirow{4}{*}{\textit{OCT}}      
    & Lee \etal \cite{Lee2017Deep} & \specialcell{VGG16 trained from scratch} & \specialcell{unmentioned} \\
    \cline{2-4}
    & Treder \etal \cite{Treder2018Automated}  & \specialcell{Fine-tuned Inception-V3} & \specialcell{$\circ$ flip} \\
    \cline{2-4}
    & Karri \etal \cite{Karri2017Transfer}  & \specialcell{Fine-tuned GoogLeNet}  & \specialcell{unmentioned} \\
    \cline{2-4} 
    & Russakoff \etal \cite{Russakoff2019Deep}  & \specialcell{A customized CNN} & \specialcell{$\circ$ rotation, additive noise}  \\
\hline
\multirow{2}{*}{\textit{\specialcell{CFP + OCT}}}      
                      & Yoo \etal \cite{YooThe}  & \specialcell{VGG19 features + random forest classifier} &\specialcell{$\circ$ translation, rotation, brightness change, additive noise} \\
                      \cline{2-4}
                      & \textit{This work} & \specialcell{Two-Stream CNN} & \specialcell{$\circ$ crop, flip, rotation, changes in  brightness / saturation / contrast\\
                       $\bullet$ \textit{CAM-conditioned image synthesis} \\ $\bullet$ \textit{Loose pairing}}  \\
 \hline
\end{tabular}
}
\end{table*}

%% file: fig-gan.tex
\begin{figure}[b!]
    \centering
    \subfigure[pix2pix]{\label{fig:pix2pix}
    \includegraphics[width=\columnwidth,center]{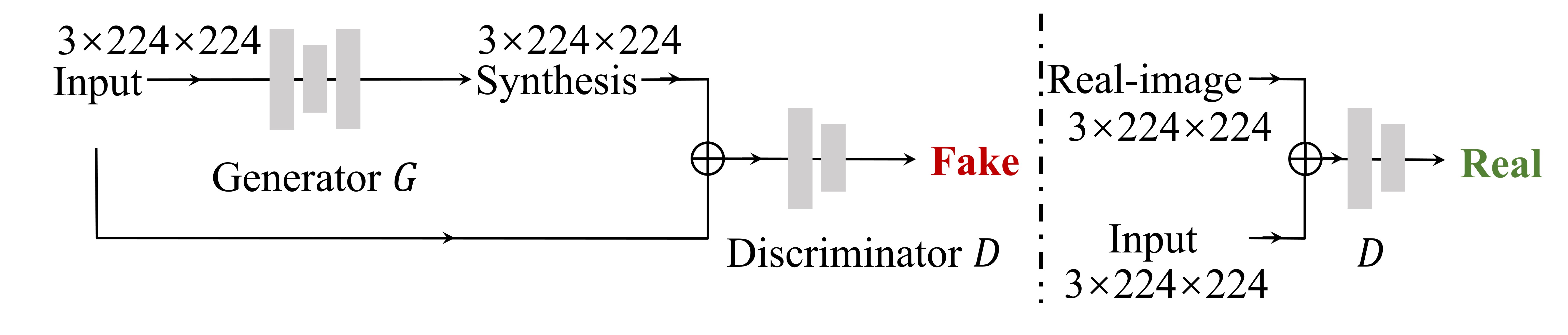}}
    \subfigure[pix2pixHD]{\label{fig:pix2pixhd}
    \includegraphics[width=\columnwidth,center]{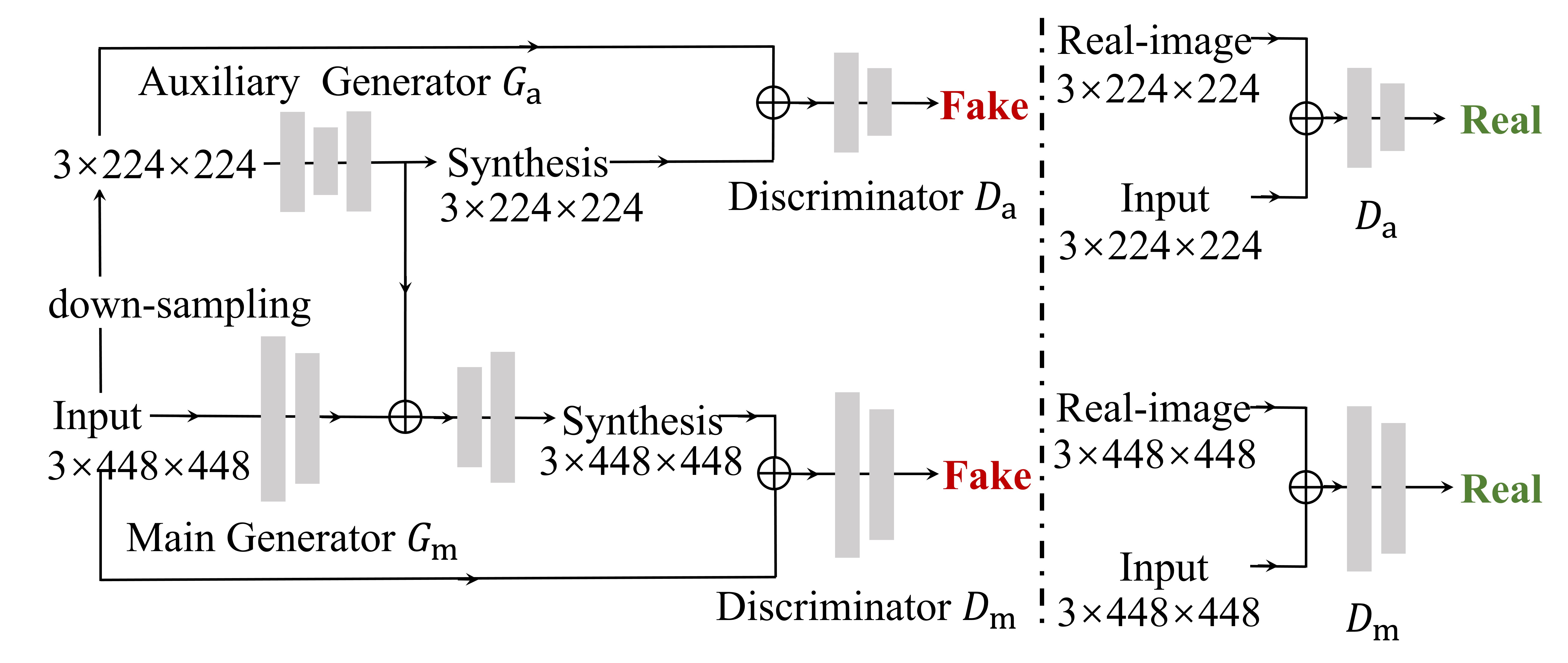}}
    \caption{\textbf{Conceptual illustrations of pix2pix and pix2pixHD for image-to-image translation}. A pix2pix model converts an input image to a synthetic image using a generator network $G$, the parameters of which are optimized during training together with  a discriminator network $D$. Note that $\oplus$ indicates channel-wise concatenation. A pix2pixHD model generates higher-resolution images in a coarse-to-fine manner. An auxiliary generator $G_{a}$ and an auxiliary discriminator $D_{a}$ are first trained for lower-resolution synthesis. Then, a main generator $G_{m}$ and a main discriminator $D_{m}$ are used to double the resolution.}
    \label{fig:p2pgan}
    \end{figure}

%% file: method.tex
%Given a color fundus image $I_f$ and an OCT image $I_o$ taken from a specific eye, we aim to build a multi-modal CNN (MM-CNN) that takes the paired input and categorizes the eye's condition to a specific class $c$:

\revise{Given a CFP image $I_f$ and an OCT image $I_o$ acquired from a specific eye}, 
our goal is  to categorize the condition of \revise{the} eye into one of the four classes, \ie $\{$\textit{normal}, \textit{dryAMD}, \textit{PCV}, \textit{wetAMD}$\}$. % given from the given eye. To that end, 
We propose a multi-modal CNN (MM-CNN) \revise{to predict} a specific class $c$ based on the paired input $\{I_f, I_o\}$.
%:
%\begin{equation} \label{eq:general}
%c \leftarrow \mbox{MM-CNN}(\{I_f, I_o\}).
%\end{equation}
%with $c \in \{normal, dryAMD, PCV, wetAMD\}$. 
%In what follows, we describe 
%
MM-CNN \revise{is described in} Sec. \ref{ssec:network}, followed by our \xre{data augmentation} methods 
%for multi-modal data augmentation 
in Sec. \ref{ssec:augmethod}. The training strategy for MM-CNN is detailed in Sec. \ref{ssec:training}.

\subsection{MM-CNN for AMD Categorization} \label{ssec:network}

\subsubsection{Network Architecture} 

%To handle the multi-modal input, we design a two-stream network as illustrated in Fig. \ref{fig:pipeline}. It consists of two symmetric branches, one for processing the fundus image $I_f$ and the other for processing the OCT image $I_o$. Note that such an architecture resembles to some extent the two-stream network widely used for video action recognition \cite{feichtenhofer2016convolutional}. The major difference is at which layer multi-modal fusion is performed. Feature maps generated by intermediate layers of a CNN preserves, to some extent, the spatial information of an input image. As different streams of video data are spatially correlated, the state-of-the-art for video action recognition performs fusion by combining feature maps from the individual streams \cite{feichtenhofer2016convolutional}. By contrast, as $I_f$ and $I_o$ are not spatially correlated, we opt to perform the fusion after the global average pooling (GAP) layer, which removes the spatial information by reducing each feature map into a single value.
We instantiate MM-CNN with a two-stream CNN. 
%\ws{A multi-modal input instance is an image pair consisting of a color fundus and an OCT image. To deal with such paired instance, we design a two-stream CNN as shown in Fig. \ref{fig:pipeline}. 
As shown in Fig. \wsrevise{\ref{fig:pipeline:a}}, the network contains two symmetric branches, which process $I_f$ and  $I_o$ in parallel. \xre{T}his type of network architecture bears some similarity to the two-stream network investigated in the context of human action recognition in videos~\cite{feichtenhofer2016convolutional}. \xre{T}he main distinction lies in at which layer multi-modal fusion is conducted. Recall that a single-modal CNN extracts information from a given image by producing arrays of 2-D feature maps layer-by-layer. While being continuously down-sized, these feature maps preserve the spatial information of the input image to some extent. As for video data, its RGB stream and Optical Flow stream are spatially correlated. It is this reason that fusion by combining feature maps from the distinct streams is preferred for video action recognition~\cite{feichtenhofer2016convolutional}. Contrary to video data, the \revise{CFP} image and the OCT image lack such spatial correlation, as noted in Sec. \ref{sec:intro}. We therefore opt for spatially invariant fusion, achieved by performing the fusion after the Global Average Pooling (GAP) layer, which compresses each feature map to a scalar and thus removes the spatial information.

%\ws{This architecture resembles to the two-stream network that has been actively investigated in the context of video action recognition \cite{feichtenhofer2016convolutional}. Considering of the spatial relationship, we make change at the fusion layer in which the features of two modalities are combined together. To be specific, feature map outputted by intermediate convolutional block contains the spatial information of the input image, while the inputs of different streams in video recognition network are spatially correlated, so the state-of-the-art for the task implements fusion operation by combining feature maps from two streams \cite{feichtenhofer2016convolutional}. By contrast, note that color fundus and OCT images are not spatially correlated, and we leverage the global average pooling (GAP) layer before fusing features to remove the spatial information by averaging all pixels in each feature map into a single value.}

%For each branch, we use convolutional blocks of ResNet-18 \cite{He2016Deep}, which has been pre-trained on the ImageNet dataset~\cite{deng2009imagenet}. In principle, any other state-of-the-art CNN can be used here. We choose ResNet-18 as it has fewer parameters and thus requires less training data. Also, this CNN is shown to be effective for other fundus image analysis tasks \cite{mmm2019-left-right-eye}. 

For each branch of the two-stream CNN, we adopt ResNet-18 as their backbone. While other CNNs can also be used, our consideration of choosing ResNet-18 is as follows. This network has relatively fewer parameters and thus needs less data for training, meanwhile its effectiveness has been justified in other fundus image recognition tasks~\cite{accv2018-laser-scar,mmm2019-left-right-eye}. Note that for the OCT image, we convert each of its pixels from grayscale to RGB by duplicating the intensity for each RGB component. \revise{The CFP and OCT images are normalized into the range  $[-1,1]$ with mean and std of $0.5$, using a default PyTorch normalization operation\footnote{torchvision.transforms.Normalize(mean=(0.5, 0.5, 0.5), std=(0.5, 0.5, 0.5))}}. As such, the same architecture and initialization are applied to both branches. 

%\ws{We leverage ResNet-18 that is pre-trained on the ImgaeNet dataset ~\cite{deng2009imagenet} as the backbone of each stream in MM-CNN. Note that any other state-of-the-art CNN is optional here. ResNet-18 is selected in this work as it relatively has fewer parameters and thus need less data in training, reducing the negative influence of lacking data to some degree. In addition, the network is shown effectiveness to process color fundus images \cite{mmm2019-left-right-eye}}

%Let $\mathbf{F}_f=\{F_{f,1},\ldots, F_{f,512}\}$ be an array of $m \times m$ feature maps generated by the ResNet-18 module in the fundus branch. The value of $m$ depends on the size of the input, which is $14$ for an input size of $448 \times 448$. Given a specific feature map $F_{f,i}$, the value of a specific position $(x,y)$ is acquired as $F_{f,i}(x,y)$. In a similar vein, we define the feature maps for the OCT branch as $\mathbf{F}_o=\{F_{o,1},\ldots, F_{o,512}\}$. 

\textbf{Spatially-invariant fusion}. For the \revise{CFP} branch, the last convolutional block of ResNet-18 produces an array of 512 feature maps, denoted as $\mathbf{F}_f = \{F_{f,1},\ldots, F_{f,512}\}$. Each feature map has a size of $m \times m$, with $m$ subject to the size of the input image. Instead of the commonly used input size of $3 \times 224 \times 224$, we choose a larger size of $3 \times 448 \times 448$, which shows better performance in our preliminary experiment\footnote{Compared with the input size of $3\times 224 \times 224$, the larger input size of $3\times 448 \times 448$ leads to loss in accuracy (from 0.721 to 0.717) for \revise{CFP-CNN}, yet improvements for both OCT-CNN (from 0.811 to 0.818) and MM-CNN (from 0.795 to 0.804).}. Accordingly, the value of $m$ is 14. For a specific feature map $F_{f,i}$, we use $F_{f,i}(x,y)$ to access the \xre{feature} value \xre{at} a specific position $(x,y)$. In a similar style, the feature maps from the OCT branch are defined as $\mathbf{F}_o=\{F_{o,1},\ldots, F_{o,512}\}$.

%\ws{We define the $m \times m$ feature maps produced by the ResNet-18 convolutional blocks in the fundus branch as . The value of $m$, which is $7$ in the original ResNet-18, is set to $14$ due to the input size of $448 \times 448$. For a specific feature map $F_{f,i}$, we denote the value of a specific position $(x,y)$ as $F_{f,i}(x,y)$. As for the OCT branch, we similarly present its output feature maps as $\mathbf{F}_o=\{F_{o,1},\ldots, F_{o,512}\}$.}

Our spatially-invariant fusion layer is \xre{simply} implemented as follows. For $\mathbf{F}_f$ and $\mathbf{F}_o$, they are fed into a GAP layer in parallel, resulting in two 512-d feature vectors, indicated by $\bar{F}_{f}=(\bar{F}_{f,1}, \ldots, \bar{F}_{f,512})$ and $\bar{F}_{o}=(\bar{F}_{o,1}, \ldots, \bar{F}_{o,512})$, respectively. A concatenation operation merges $\bar{F}_{f}$ and $\bar{F}_{o}$ into a new 1024-d vector with semantic information from both modalities \xre{combined}. For \xre{the} four-class classification \xre{task}, by feeding the merged vector into a fully connected layer of shape $1024 \times 4$, we obtain $s^c$ as a decision score for a specific class:
%
%\ws{$\mathbf{F}_f$ and $\mathbf{F}_o$ are the separate features extracted from individual modalities, and a fusion layer is performed to combine them. The multi-modal feature maps are first fed in to a GAP layer, obtaining two $1\times 512$ vectors that are respectively defined as $\bar{F}_{f}=(\bar{F}_{f,1}, \ldots, \bar{F}_{f,512})$ and $\bar{F}_{o}=(\bar{F}_{o,1}, \ldots, \bar{F}_{o,512})$. The two vectors are then concatenated into a vector in size of $1 \times 1024$, followed a fully connected (FC) layer and a softmax layer that receive the fused vector and give a final score $s^c$ for a specific class $c$, }
%
%Our fusion layer is implemented by first feeding separately $\mathbf{F}_f$ and $\mathbf{F}_o$ into a GAP layer to obtain two $1\times 512$ vectors, denoted as $\bar{F}_{f}=(\bar{F}_{f,1}, \ldots, \bar{F}_{f,512})$ and $\bar{F}_{o}=(\bar{F}_{o,1}, \ldots, \bar{F}_{o,512})$, respectively. The two vectors are then concatenated to form a $1 \times 1024$ vector which contains information from the two modalities. For classification, the combined vector is fed into a fully connected (FC) layer to produce a score for a specific class $c$, denoted as $s^c$, 
\begin{equation} \label{eq:raw-score}
s^c = \underbrace{\sum^{512}_{i=1} w^c_{f,i} \cdot \bar{F}_{f,i}}_{s^c_f} + \underbrace{\sum^{512}_{i=1} w^c_{o,i} \cdot \bar{F}_{o,i}}_{s^c_o},
\end{equation}
where $\{w^c_{f,1},\ldots,w^c_{f,512}\}$ and $\{w^c_{o,1},\ldots,w^c_{o,512}\}$ denote class-specific parameters of the fully connected layer, while $s^c_f$ and $s^c_o$ indicate decision scores contributed by the \revise{CFP} branch and the OCT branch, respectively. Classification %defined in (\ref{eq:general}) 
is realized by selecting the class maximizing $s^c$.

%\begin{equation} \label{eq:maxscore}
%c^{*} = \arg \mathop{\max}_{c} s^c.
%\end{equation}

\subsubsection{Multi-Modal Class Activation Mapping for Visual Interpretation}  \label{sssec:mm-cam}

%As \ref{eq:raw-score} shows, the classification score $s^c$ for a given class $c$ is additively contributed by both modalities. For a more intuitive interpretation, we leverage class activation mapping (CAM) \cite{Zhou2015Learning}, which reveals the (implicit) attention of a CNN on an input image. We compute the multi-modal version of CAMs as 
We note from (\ref{eq:raw-score}) that the class-specific score $s^c$ is contributed by the two modalities in an additive manner. In order to visualize their contributions, we employ Class Activation Mapping (CAM) \cite{Zhou2015Learning}. As a response-based visualization technique, CAM projects the classification score back to a $m\times m$ heatmap, and thus reveals which part of the input image contributes the most. 
\xre{To make the paper more self-contained}, 
%In what follows, 
we first depict CAM developed initially for a single-modal CNN, with \revise{CFP} images as a showcase. We then extend CAM for the multi-modal scenario.

\begin{figure}[htb!]
\centering
\begin{minipage}[b]{0.31\columnwidth}
\subfigure[Multi-modal input]{\label{fig:instance:a}
\includegraphics[width=\columnwidth]{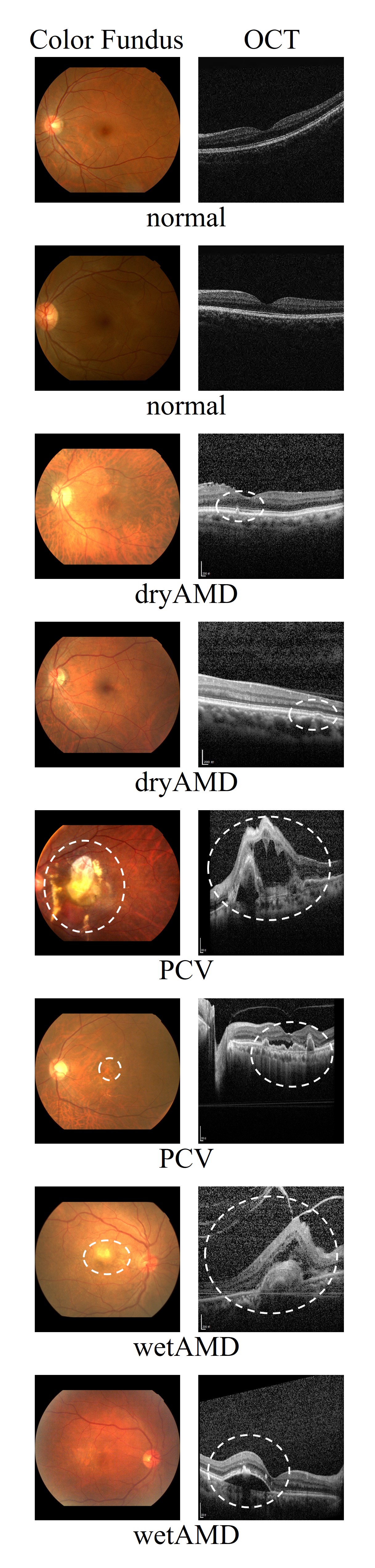}}
\end{minipage}
\begin{minipage}[b]{0.31\columnwidth}
\subfigure[Single-modal CAMs]{\label{fig:instance:b}
\includegraphics[width=\columnwidth]{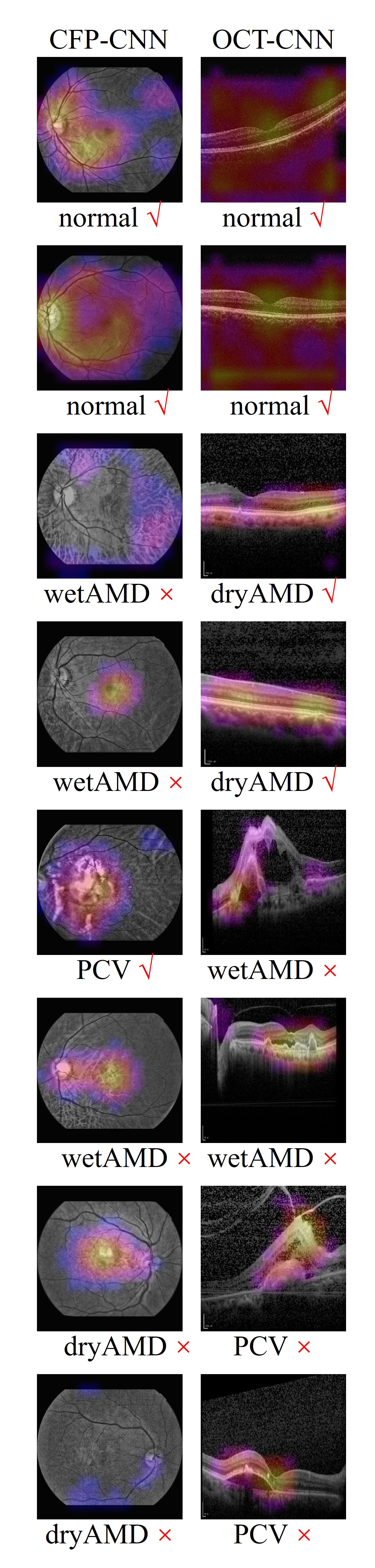}}
\end{minipage}
\begin{minipage}[b]{0.31\columnwidth}
\subfigure[Multi-modal CAMs]{\label{fig:instance:c}
\includegraphics[width=\columnwidth]{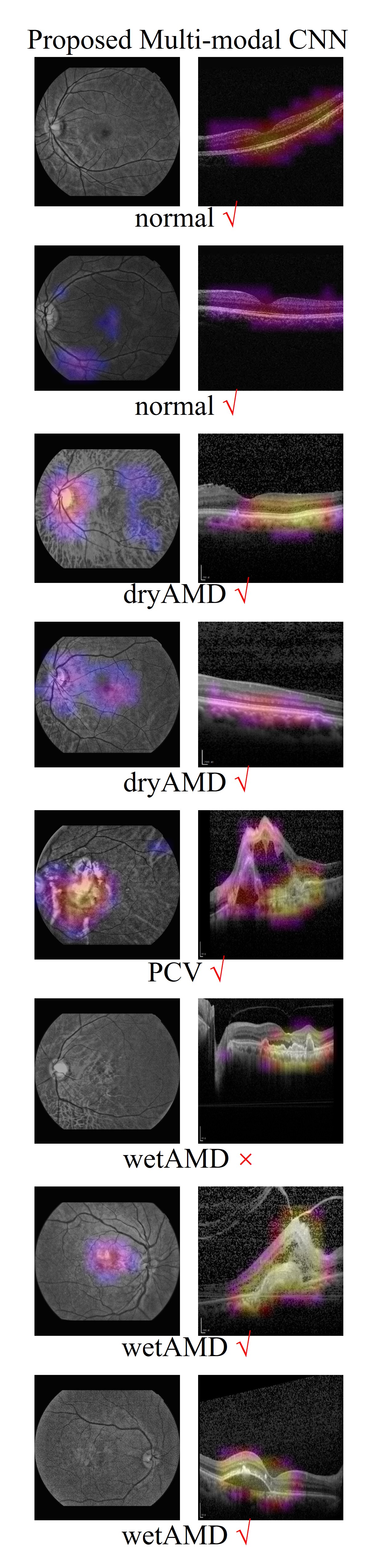}}
\end{minipage}
\caption{\textbf{CAM-based visualization of which part of input images contribute to the final predictions made by single-modal / multi-modal CNNs}.  Input images are shown in (a), with AMD-related regions marked out by dotted ellipses. In (a), labels under each pair of images are ground-truth, while labels in (b) and (c) are predictions made by corresponding models. Correct and incorrect predictions are marked out by \checkmark and $\times$, respectively. For a better visualization, the images are overlaid with single-modal CAMs (b) and multi-modal CAMs (c), whilst the \revise{CFP} images are converted to gray scale. 
%Some original images with the ground truth labels are shown in (a). We emphasize AMD-related abnormalities by the white ellipses. (b) and (c) give the predicted label of images in (a) and their corresponding visual interpretations, in which brighter color presents higher activations. For better visualizing the heat map, we transform the color fundus images from RGB to grayscale in (b) and (c).
}
\label{fig:instance}
\end{figure}

To simplify our notation, we re-purpose $s^c_f$ from (\ref{eq:raw-score}) to indicate the decision score of class $c$ made by a single-modal CNN with respect to a test \revise{CFP} image $I_f$. Recalling that $\bar{F}_{f,i}$ is equivalent to $\frac{1}{m^2} \sum_{x,y} F_{f,i}(x,y)$, we express $s^c_f$ as 
\begin{equation} \label{eq:raw-score-fundus}
\begin{array}{ll}
s^c_f & = \sum^{512}_{i=1} w^c_{f,i} \cdot \bar{F}_{f,i} \\
      & = \sum^{512}_{i=1} w^c_{f,i} \cdot (\frac{1}{m^2}\sum_{x,y} \cdot F_{f,i}(x,y) ) \\
      & = \sum_{x,y} \frac{1}{m^2} \sum^{512}_{i=1} w^c_{f,i} \cdot F_{f,i}(x,y).
\end{array}
\end{equation}
By defining $CAM^c_f$ as a $m\times m$ \xre{feature} map %with its each position 
computed \xre{by}
\begin{equation} \label{eq:single-modal-cam}
CAM^c_f(x,y) = \frac{1}{m^2} \sum^{512}_{i=1} w^c_{f,i} \cdot F_{f,i}(x,y),
\end{equation}
$s^c_f$ is equivalently the sum of $CAM^c_f$, \ie
\begin{equation} \label{eq:single-modal-score}
s^c_f = \sum_{x,y} CAM^c_f(x,y).
\end{equation}
We see from (\ref{eq:single-modal-score}) that the classification score is spatially distributed to each position the CAM map. 
Consequently, by overlaying an input image with its corresponding (up-sampled) CAM map, salient regions are highlighted, see Fig. \ref{fig:instance:b}.

%Consequently, by up-sampling the map to the size of the input image, salient regions in the input image can be highlighted, 

We derive a multi-modal version of CAM for class $c$ as 
%
%\ws{As (\ref{eq:raw-score}) shows, the final score $s^c$ for specific class $c$ is additively contributed by both modalities. To qualitatively interpret how much the individual modalities contribute, we employ class activation mapping (CAM) \cite{Zhou2015Learning} in multi-modal version, computed as (\ref{eq:cam}) ,showing the (implicit) attention of two-stream CNN on the input pair.}
%
\begin{equation}\label{eq:cam}
\left\{
\begin{array}{lr}
CAM^{c}_{f}(x,y)=\frac{1}{m^2}\sum^{512}_{i=1} w^{c}_{f,i} \cdot F_{f,i}(x,y),\\
\\
CAM^{c}_{o}(x,y)=\frac{1}{m^2}\sum^{512}_{i=1} w^{c}_{o,i} \cdot F_{o,i}(x,y).
\end{array}
\right.
\end{equation}

%Note that $\bar{F}_{f,i}$ used in (\ref{eq:raw-score}) is equivalent to $ \sum_{x,y} F_{f,i}(x,y)$, while $ \bar{F}_{f,o} $ is equivalent to $ \sum_{x,y} F_{f,o}(x,y) $. 
Putting (\ref{eq:raw-score}), (\ref{eq:raw-score-fundus}) and (\ref{eq:cam}) together, the score $s^c$ for class $c$ can be reformulated using the \wsrevise{CFP} and OCT CAM\xre{s}, % maps, 
namely
\begin{equation} \label{eq:sc-cam}
s^c = \sum_{x,y} CAM^{c}_{f}(x,y) + \sum_{x,y} CAM^{c}_{o}(x,y).
\end{equation}
%According to \ref{eq:sc-cam}, $CAM^{c}_{f}(x,y)$ and $CAM^{c}_{o}(x,y)$ indicate the contribution of a specific position of the fundus and OCT images, respectively. Consequently, we visualize the contribution of each modality by overlaying with the corresponding up-sampled CAM, see Fig. \ref{fig:instance}.
%As (\ref{eq:sc-cam}) shows, 
\xre{T}he decision score is now  distributed  \xre{via}
%in the form of 
$CAM^{c}_{f}(x,y)$ and $CAM^{c}_{o}(x,y)$ to each position \xre{on} the \revise{CFP} and OCT images, in a down-sampled size of $m \times m$. This enables us to visualize the contribution of each modality, see Fig. \ref{fig:instance}. 
%We refer to Fig. \ref{fig:instance} for real-case visualization.

%in  where we overlay the original images with the corresponding up-sampled CAM for clearer observation.}

%\input{images/fig-cam}

\subsection{Multi-Modal Data Augmentation} \label{ssec:augmethod}

\input{method-data-aug}

\subsection{A Two-stage Training Strategy for Multi-Modal CNN} \label{ssec:training}

Given a relatively small real-world dataset, the joint use of the proposed CAM-conditioned image synthesis and loose pairing enables us to construct a large amount of multi-modal training instances. However, directly combining the real dataset and the generated dataset is problematic, as the former will be easily outnumbered. In that regard, we train the MM-CNN in two stages, \ie pre-training and fine-tuning. In the first stage, a \revise{CFP}  image and an OCT image are loosely paired, with at least one of them sampled from the generated dataset. Given the MM-CNN trained on these fake pairs, we perform fine-tuning in the second stage, using the real dataset.
%and again with loose pairing. 
Despite the simplicity, such a two-stage training strategy is found to be effective, as we will shortly show in Sec. \ref{ssec:exp2}.

\xre{F}or the backbone network, we start with  ResNet-18  pre-trained on ImageNet~\cite{deng2009imagenet}. \xre{Instead of using} the original input size is $3 \times 224 \times 224$, we use a larger input \xre{size} of $3 \times 448 \times 448$. The kernel size of the GAP layer is accordingly adjusted, from $7 \times 7$ to $14 \times 14$, to ensure the dimensionality of the last feature vector is invariant to this change. For both single-modal and multi-modal CNNs, we use cross-entropy, a common loss function for multi-class classification. SGD is used as the optimizer, with momentum of 0.9, weight decay of 1e-4 and batch size of 8. Per training process, a model is selected based on its performance on a held-out validation set.

%\ws{Referring to \cite{Jintasuttisak2014Color}, 
Contrast-limited adaptive histogram equalization (CLAHE) \cite{Jintasuttisak2014Color} is used to enhance CFP images. 
%see the Appendix for its effect on the performance. 
For noise reduction in OCT images, median filtering with a $3\times 3$ kernel is applied. \revise{Note that the proposed data augmentation methods are not meant for replacing conventional data augmentation strategies. In fact,  the conventional strategies are needed to train better single-modal CNNs for \xre{CAM-based} image synthesis. The proposed strategies and the conventional strategies shall be used in combination.} In advance to the proposed data augmentation, we perform low-level common data augmentation operations including random crop, flip, rotation, and random changes in contrast, saturation and brightness on training images.

%% file: method-data-aug.tex
For effectively increasing the amount of multi-modal training instances, we propose two data augmentation methods, one for image-level (Sec. \ref{sssec:gan}) and the other for pair-level (Sec. \ref{sssec:loose-pairing}). Both methods are applied only on the training data.

\subsubsection{CAM-conditioned Image Synthesis for Image-level Data Augmentation} \label{sssec:gan}

In order to generate \wsrevise{CFP} / OCT images for a specific AMD class $c$, we re-purpose CAMs as input of pix2pixHD~\cite{wang2018high}, a powerful network for image-to-image translation. The proposed CAM-conditioned image synthesis \xre{method} is illustrated in Fig. \wsrevise{\ref{fig:pipeline:b}}.

\begin{figure}[htb!]
\centering
\subfigure[Real CFP images, CAMs, and synthesized CFP images]{
\includegraphics[width=\columnwidth]{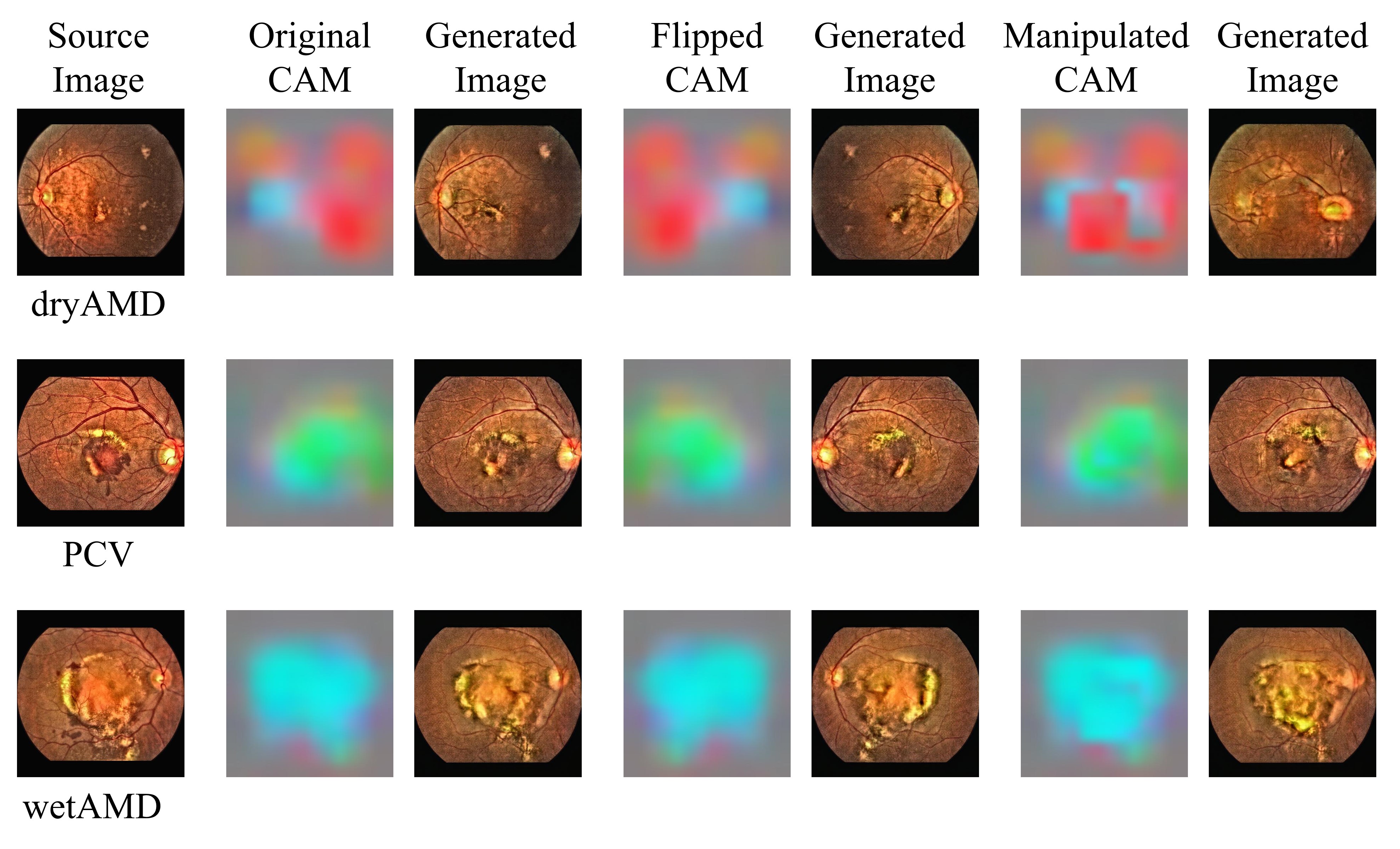}
\label{fig:syn-cfp}}
\subfigure[Real OCT images, CAMs, and synthesized OCT images]{
\includegraphics[width=\columnwidth]{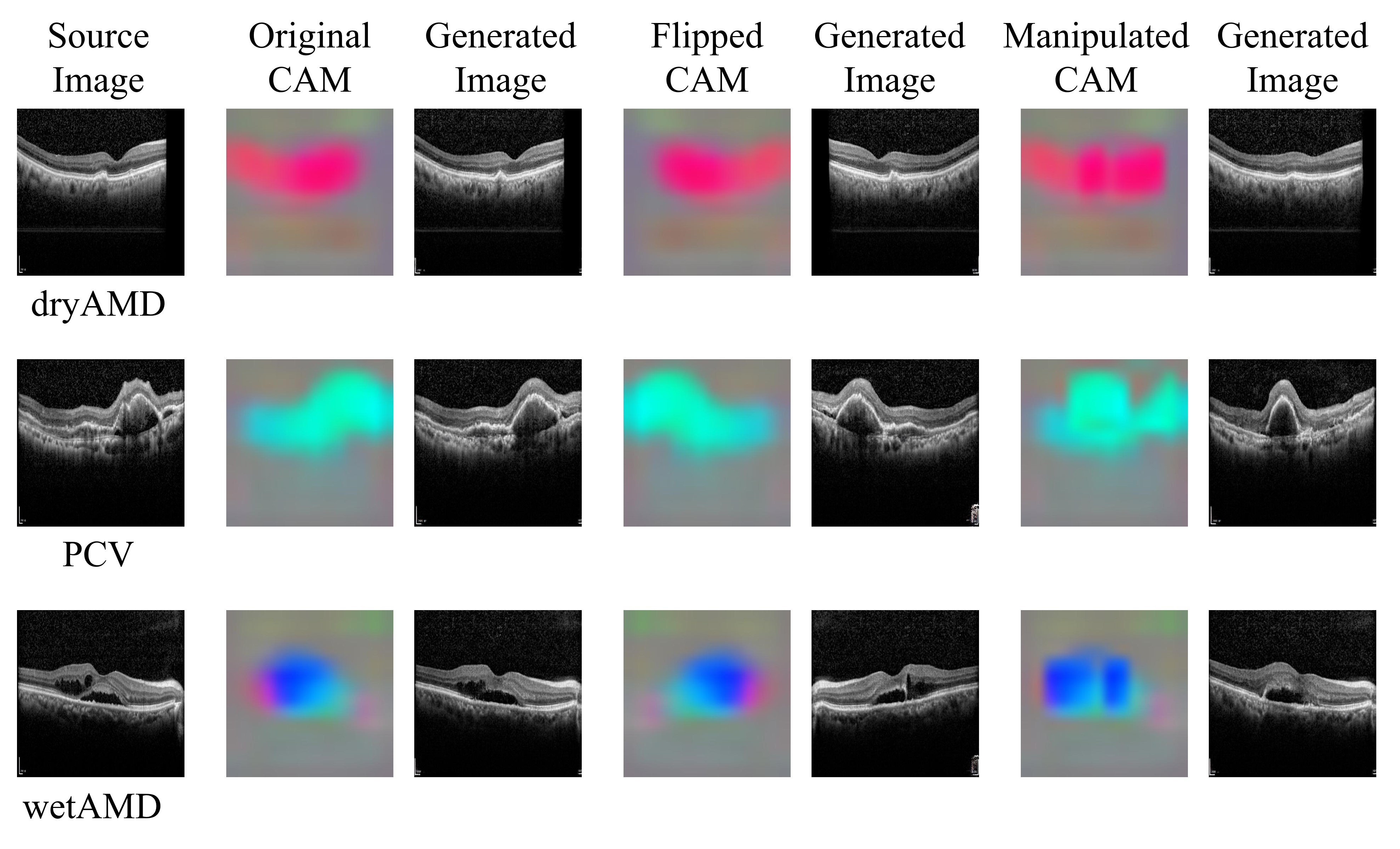}}
\caption{\textbf{Illustrations of images generated by the CAM-conditioned image synthesis} for (a) CFP and (b) OCT images. For better visualization, for each source image we stack its three CAMs, \ie $CAM^{dry}$, $CAM^{pcv}$ and $CAM^{wet}$ to form an RGB image. The fact that dominant colors are better observed in OCT images suggests this modality is more suited for AMD categorization. A flipped CAM indicates horizontal flip of an original CAM, while a manipulated CAM is obtained by moving the highest response region. Best viewed in digit.}
\label{fig:syn}
\end{figure}

\textbf{From a source image to CAMs}. 
Given a source image $I$ from a specific AMD class $c$, we first use a single-modal CNN, which we have pre-trained, % for AMD categorization, 
to produce CAMs \textit{w.r.t.} the three AMD classes. We ignore the normal class, as \xre{its} CAMs do not reflect abnormalities by definition. %, so we ignore it.
As described in Sec. \ref{sssec:mm-cam}, for an input image of $3\times 448 \times 448$, each CAM computed by (\ref{eq:single-modal-cam}) has a size of $14\times 14$. The CAMs are stacked to form a three-channel image $I_{cam}=[CAM^{dry}; CAM^{pcv}; CAM^{wet}]$. With the CAMs treated as red, green and blue channels in an RGB image, $I_{cam}$ is colorized naturally. As manifested by the second column of Fig. \ref{fig:syn}, dominant colors are better observed in CAMs of OCT images, suggesting this modality is more suited for AMD categorization. Next, we describe how to convert $I_{cam}$ to a positive instance \textit{w.r.t.} the given class. 

%\begin{figure}[tbh!]
%\centering
%\subfigure[]{
%\label{fig:instance:a}
%\includegraphics[width=\columnwidth]{images/gan-f}}
%\subfigure[]{
%\label{fig:instance:b}
%\includegraphics[width=\columnwidth]{images/gan-o}}
%\caption{\textbf{CAM-conditioned fundus / OCT image synthesis}. Given a source image of $3 \times 448 \times 448$, let it be (a) fundus or (b) OCT, we use the corresponding single-modal CNN to produce CAMs with respect to each AMD class. The CAMs are stacked to form an three-channel image $[CAM^{dry}; CAM^{pcv}; CAM^{wet}]$ of $3 \times 14 \times 14$, which is then fed into an image-to-image translation GAN \cite{wang2018high} for image synthesis. Manipulating the CAMs results in multiple synthesized images.}
%\label{fig:cgan}
%\end{figure}

\textbf{From CAMs to synthesized images}. 
%Different from a classical GAN that has one generator and one discriminator, pix2pixHD consists of a main generator $G_m$ and an auxiliary generator $G_a$ that produces images at two resolutions, which are $3 \times 448 \times 448$ and $3 \times 224 \times 224$ in this work. Accordingly, there are two discriminators responsible for the two resolutions. Note that we do not draw the discriminators in Fig. \ref{fig:pipeline} as they are used only in the training stage. A new image is generated by $G_m$ with assistance from $G_a$. In our context, $I_{cam}$ is first up-sampled by bicubic interpolation to two scales, \ie $3 \times 224 \times 224$ and $3 \times 448 \times 448$. The two enlarged CAMs are separately fed to $G_a$ and $G_m$. As illustrated in Fig. \ref{fig:pipeline}, information extracted by $G_a$ is injected into $G_m$ by combing their feature maps by the matrix addition. For each modality, we train a pix2pixHD model, following the training procedure as described in~\cite{wang2018high}. \ws{Horizontal flip with probability 0.5 and random crop are utilized as data augmentation for this GAN.}
%We adopt pix2pixHD for color fundus / OCT image synthesis. 
Different from a classical GAN that has one generator and one discriminator, pix2pixHD consists of a main generator $G_m$ and an auxiliary generator $G_a$ that produce 
%designed in U-net framework~\cite{olaf2015unet} which is made up of a down-sampling block, a sequence of convolution blocks, and an up-sampling block. $G_m$ and $G_a$ produce 
images at two different resolutions, which are $3 \times 448 \times 448$ and $3 \times 224 \times 224$ in this work, see Fig. \ref{fig:pix2pixhd}. Accordingly, there are two fully convolutional network based discriminators $D_m$ and $D_a$ responsible for the two resolutions. Note that we do not draw the discriminators in Fig. \wsrevise{\ref{fig:pipeline:b}} as they are used only in the training stage. A new image is generated by $G_m$ with assistance from $G_a$. 
In our context, $I_{cam}$ is first up-sampled by bicubic interpolation to two scales, \ie $3 \times 224 \times 224$ and $3 \times 448 \times 448$. The two enlarged CAMs are separately fed to $G_a$ and $G_m$. As illustrated in Fig. \ref{fig:pix2pixhd}, information extracted by $G_a$ is injected into $G_m$ by combing their feature maps by the matrix addition. For each modality, we train a pix2pixHD model, see Section \ref{ssec:expset} for details of the training procedure.

We observe that the highest response region in the CAM tends to indicate abnormal areas in the source image. Hence, in order to synthesize diverse yet meaningful images, we manipulate $I_{cam}$ by moving around its highest response region. Specifically, we localize such a region by a sliding-window approach, with the window size empirically set to be $5 \times 5$. The window that has the maximal response with respect to the given class is selected. Then, the region is \revise{swapped} with a randomly chosen region of the same size. In this manner we construct for the same source image a number of distinct CAMs, which we term \textit{manipulated CAMs}, to distinguish them from the original CAM. Fig. \ref{fig:syn} showcases some source images, their original and manipulated CAMs, and synthesized images. 

When dealing with a novel image, it is possible that the CAM may mis-localize some AMD lesions and thus result in a wrong CFP / OCT image. To ensure the credibility of CAM visualization, our tactic here is to re-use the training images that are already been seen by CFP-CNN / OCT-CNN in their training stage. Such an ``overfitting'' produces good-quality CAM in general. In fact, we observe that feeding a CAM to pix2pixHD reconstructs its source image well. Manipulating the CAM by swapping is mainly for diversifying the synthetic data. Moreover, the synthetic data will be used exclusively for pre-training MM-CNN. We expect to reduce the negative effect of wrongly generated images by fine-tuning.

%Different colors of CAMs represent the activation of different categories. The class label of a synthetic image depends on the dominant color in its corresponding CAM.}

%\ws{Unfortunately, not all synthetic images are of high quality. Fig. \ref{fig:gancmp:b} shows some images with artifact. Note that these low-quality images are not manually filtered out and also used for training MM-CNN.}

%\begin{figure}[tbh!]
%\centering
%\includegraphics[width=\columnwidth]{images/bad_synthesis.jpg}
%\caption{\textbf{Low-quality synthetic images}. \ws{Dotted ellipses mark out the visually unrealistic regions. The image (f-a) seems to have two optic disks, while (f-b) has no optic disc structure. The images (f-a), (f-b), and (f-c) have rough boundaries. Also in (f-c), there is a black region indicating failed synthesis. As for OCT synthesis, the images (o-a), (o-b) and (o-c) show no information about layering that OCT images should have. Moreover, the images (f-d), (o-d) and (o-e) include obvious artifacts. Best viewed in digit.}}
%\label{fig:bad}
%\end{figure}

\subsubsection{Loose Pairing for Pair-Level Data Augmentation} \label{sssec:loose-pairing} 
To obtain a multi-modal training example, a natural strategy is to strictly select a CFP image and an OCT image based on their eye identities. Suppose we have two PCV eyes, $eye^{a}$ and $eye^{b}$, in the training set. Each eye is associated with several images, \ie $set^a = \{f^{a},o^{a}_{1},o^{a}_{2},o^{a}_{3}\} $ and $set^b = \{f^{b},o^{b}_{1},o^{b}_{2}\}$, where $f$ and $o$ stand for CFP and OCT images, respectively. There are five strict pairs only, \ie $ \{(f^{a},o^{a}_{i}),(f^{b},o^{b}_{j})\}$, $i=1,2,3$ and $j=1,2$. In order to increase the number of multi-modal instances for training, we propose to construct input pairs based on \textit{labels} instead of eyes. A color fundus image can be paired with an OCT image as long as their class labels are identical. We coin this method \textit{Loose Pairing}. Consequently, we obtain five loose pairs from $set^a$ and $set^b$ and ten multi-modal instances in total. With loose pairing, the size of the training set is substantially increased.

%% file: eval.tex
\subsection{Experimental Setup}\label{ssec:expset}

\textbf{Multi-modal dataset}. We collected our experimental data from the outpatient clinic of the Dept. of Ophthalmology, Peking Union Medical College Hospital. The dataset initially contains 1,09\revise{4} \revise{CFP} images from 1,09\revise{3} eyes of \revise{829 subjects}, acquired by a Topcon fundus camera. For 8\revise{17} eyes, they are associated with one to five OCT images. The OCT images are central B-scans acquired by a Topcon OCT camera and a Heidelberg OCT camera and manually selected by technicians. 
We chose these cameras as they were the most frequently used in our outpatient clinic, allowing us to collect a decent amount of samples for this research. The condition of each eye is jointly assessed by two ophthalmologists based on its \revise{CFP}, OCT, Fluorescein angiography (FA) or indocyanine green angiography (ICGA), when applicable. Accordingly, each eye is categorized into \textit{normal}, \textit{dryAMD}, \textit{PCV} or \textit{wetAMD}. Images associated with a specific eye gets the same label. 

Eventually, we obtain an expert-labeled multi-modal dataset of 1,09\revise{4} \revise{CFP} and 1,2\revise{89} OCT images. All these data are approved by the IRB of Peking Union Medical College Hospital. We obey the principles of the Declaration of Helsinki.

\input{data_tab}

In order to construct a class-balanced test set,
%for multi-modal AMD categorization, 
for each class we randomly sample 20 eyes from the eyes that have both \revise{CFP} and OCT images. This setting enables us to compare an multi-modal input with its single-modal counterpart. The setting also allows us to make a head-to-head comparison between the two single modalities, namely \revise{CFP} versus OCT. In a similar manner, we build a multi-modal validation set for model selection. We use the remainder for training. Data split is based on eye identities, so images from a specific eye do not appear in more than one subset. 

Note that the relatively small amount of training samples for dryAMD will adversely affect the stability of \revise{CFP} image based models. Its impact on OCT-based models is minor, as dryAMD-related visual patterns such as drusen are easily recognizable in OCT images. The effect of data shuffle on the model performance is provided in the Appendix.

\textbf{Performance metrics}. For class-specific evaluation, we report \textit{Sensitivity}, \textit{Specificity} and the \textit{F1} score, which is the harmonic mean between Sensitivity and Specificity. For overall performance evaluation, we report F1 averaged over the four classes. Besides, we report \textit{Accuracy}, the ratio of correctly categorized instances, which are \revise{CFP} or OCT images for single-modal CNNs and \wsrevise{CFP}-OCT pairs for MM-CNNs. Due to randomness in SGD based training, the performance of a model obtained in a specific training round may vary. To reduce such a random effect, for each model we run training and evaluation three times, and report averaged score per metric. That is, for a given model, its reported Sensitive, Specificity and F1 scores per class are all averaged based on the three runs. Hence, the reported F1 is not the harmonic mean between the reported Sensitivity and Specificity.

\textbf{Implementation}. For fair comparisons, all CNN models assessed in our experiments use ResNet-18, pre-trained on ImageNet, as their backbones. For all the models, common data augmentation is applied for training. A model trained with the proposed data augmentation methods, when applicable, will be postfixed with \textit{-da}. \revise{Per abnormal image in the original training data, we synthesize three new images, yielding 201 dryAMD / 777 PCV / 1,359 wetAMD for the CFP modality and 99 dryAMD / 867 PCV / 1,593 wetAMD for the OCT modality. Loose pairing these synthetic images results in nearly 2.9M multi-modal samples for pre-training. As for finetuning, loose pairing is conducted on the real set, resulting in 317.6K multi-modal samples.} All deep models are implemented using PyTorch ~\cite{paszke2019pytorch} on a Ubuntu 16.04 server with an
\revise{NVIDIA® GeForce® RTX 2080 Ti} GPU.

For the single-modal baselines, two ResNet-18 models are separately trained on the \revise{CFP} and OCT images, referred to as \revise{CFP}-CNN and OCT-CNN. For both models, their input dimension is $3 \times 448 \times 448$. They are trained in a similar manner as that of MM-CNN described in Section \ref{ssec:training}, \ie SGD as the optimizer, momentum of 0.9, weight decay of 1e-4 and mini-batch size of 8. \revise{To deal with the imbalanced training samples, per mini-batch we select samples of the four classes in equal proportions.}

Following~\cite{wang2018high}, we train pix2pixHD in a coarse-to-fine manner. Specifically, for the first $p$ epochs, only the auxiliary generator $G_a$ and the auxiliary discriminate $D_a$ are trained to generate images of $3 \times 224 \times 224$. For the next $q$ epochs, the main and auxiliary networks are jointly trained to synthesize images of  $3 \times 448 \times 448$. Horizontal flip with probability of 0.5 and random crop are utilized as data augmentation. Training a pix2pixHD model on our dataset (batch size $1$) typically requires 150 epochs, with $p=100$ and $q=50$. This amounts to a training time of around 8 hours, exploiting about 120k augmented images in total. We did not observe model collapse. Artifacts, \eg synthesized \revise{CFP} images with two optic discs, are occasionally observed, see Fig. \ref{fig:gancmp}.

\subsection{Experiment 1. Multi-modal \textit{versus} Single-modal} \label{ssec:exp1}

\input{eval-single-vs-mm}

\subsection{Experiment 2. Ablation Study on Data Augmentation} \label{ssec:exp2}
\input{eval-da}

\subsection{Experiment 3. Comparison to the State-of-the-Art} \label{ssec:exp3}
\input{eval-sota}

\subsection{Discussion} \label{ssec:discussion}
\input{discussion}

%% file: data_tab.tex
\begin{table}[tbh!]
    \renewcommand{\arraystretch}{1.2}
    \centering
    \caption{\textbf{Data used in our experiments}. Data split is made based on eyes. In parentheses are number of eyes per class in each split. As the number of OCT images varies per eye, OCT images of the four classes are more unbalanced. }\label{tab:data}
    \scalebox{0.9}{
    \begin{tabular}{|l | r|r | r|r | r|r |}
    \hline
       
    \multirow{2}{*}{\textbf{Class}} & \multicolumn{2}{c|}{\textbf{Training set}} & \multicolumn{2}{c|}{\textbf{Validation set}}  & \multicolumn{2}{c|}{\textbf{Test set}}  \\
     \cline{2-7}
    
    & \textit{\wsrevise{CFP}} 	& \textit{OCT} 	& \textit{\wsrevise{CFP}} 	& \textit{OCT}	& \textit{\wsrevise{CFP}} 	&  \textit{OCT} \\
    \hline
    \textit{normal} 	& 155 (155) 		& 156 (155)		& 20 (20) 			& 20 (20)  	& 20 (20)			& 20 (20) \\
    \hline
    \textit{dryAMD} 	& 67 (~67) 		& 33 (~22) 		& 20 (20)			& 35 (20)  	& 20 (20)			& 38 (20) \\
    \hline
    \textit{PCV} 	& 2\revise{59} (2\revise{59}) & 289 (15\revise{6}) & 20 (20) & 44 (20)  & 20 (20)	& 47 (20) \\
    \hline
    \textit{wetAMD} 	& 45\revise{3} (45\revise{2}) 		& 53\revise{1} (32\revise{5}) 		& 20 (20)			& 38 (20)  	& 20 (20)			& 38 (20) \\
    \hline
    \end{tabular}
    }
    \end{table}

%% file: eval-single-vs-mm.tex
\textbf{Baselines}. Our single-modal baselines are \revise{CFP}-CNN and OCT-CNN. To study if the synthetic images help the single-modal models, we train them using the two-stage strategy described in Sec. \ref{ssec:training}. The resultant models are named as \revise{CFP-CNN}-\textit{da} and OCT-CNN-\textit{da}. As for multi-modal baselines, we consider the following three methods, \ie EarlyFusion, LateFusion and MM-MHSA. EarlyFusion stacks the CFP and OCT images into a 4-channel input, while LateFusion combines CFP-CNN and OCT-CNN by averaging the output of their softmax layers. MM-MHSA uses  multi-modal multi-head self attention \cite{yu2019multimodal} to fuse CFP and OCT features\footnote{MM-MHSA is implemented as follows. Given $\bar{F}_f$ and  $\bar{F}_o$ produced by CFP-CNN and OCT-CNN, each feature first goes through a specific $512\times 512$ FC layer and is stacked afterwards to form a $2\times 512$ feature array. The feature array is then fed into a 4-head self-attention block followed by an Add\&Norm layer, a feedforward network and another Add\&Norm. The resultant $2\times 512$ features are flattened and fed to a $1024\times 4$ FC layer for classification.}.

\textbf{Results}. 
As Table \ref{tab:s_v_m} shows, MM-CNN, trained without using the proposed data augmentation, is less effective than OCT-CNN (0.872 \textit{vs} 0.886 in F1 and 0.804 \textit{vs} 0.818 in Accuracy). Similarly, EarlyFusion, LateFusion and MM-MHSA do not surpass OCT-CNN. These results suggest that exploit two modalities does not necessarily guarantee better performance. 

%For learning an effective two-stream CNN, more training data is required. 

\input{table_result_s_vs_m}

\begin{figure}[htb!]
\centering
\includegraphics[width=0.95\columnwidth]{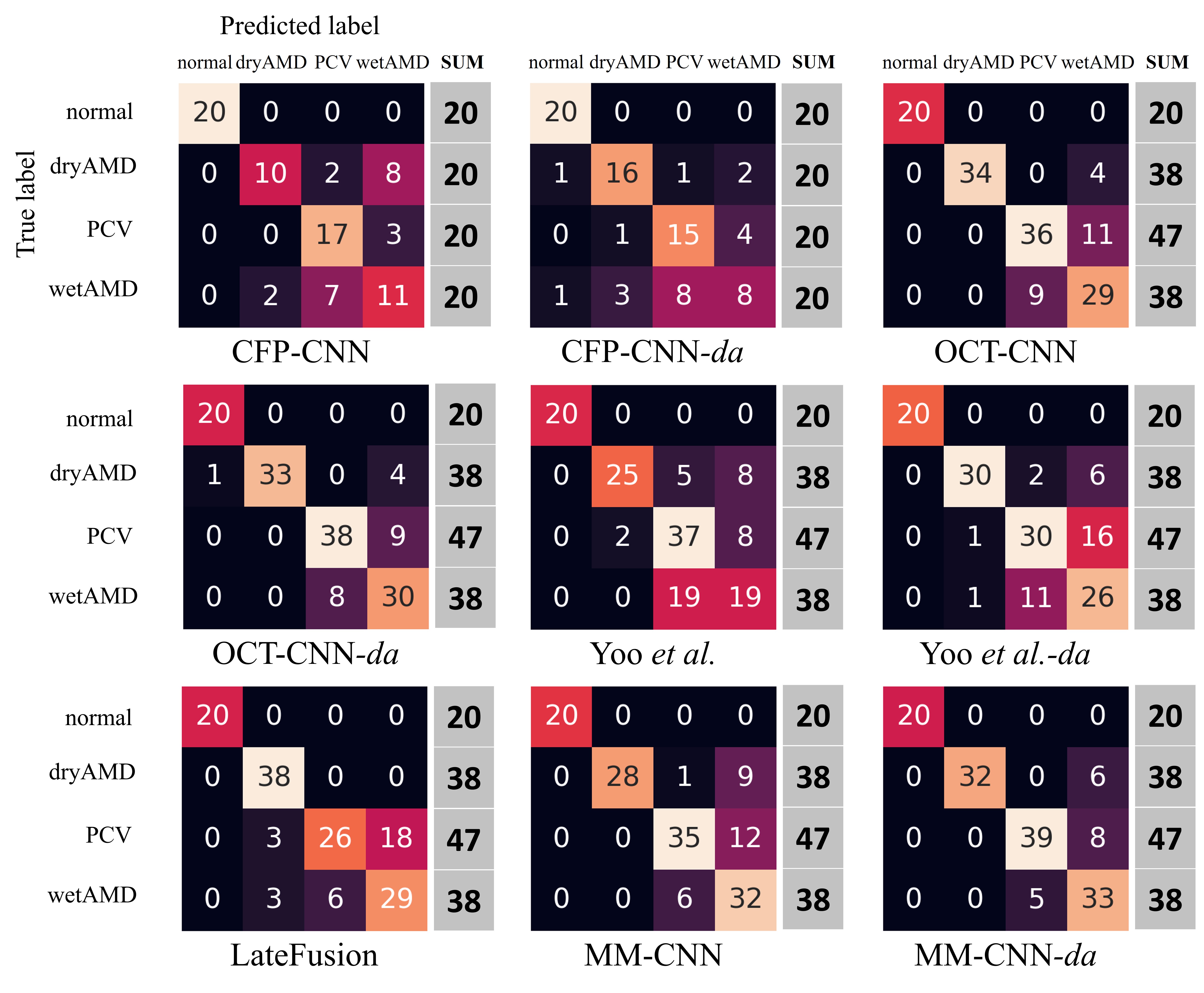}
\caption{\textbf{Confusion matrices}. Recall that per model we run training and evaluation three times. So for a fair comparison, we plot the confusion matrix of each model with its median-performance run. Compared to the best single-modal baseline (OCT-CNN), MM-CNN-\textit{da} effectively reduces confusion between PCV and wetAMD.}
\label{fig:cm}
\end{figure}

Compared to OCT-CNN, our  MM-CNN-\textit{da} obtains relative improvements of 3.2\% in F1 and 5.5\% in Accuracy. As Fig. \ref{fig:cm} shows, MM-CNN-da effectively reduces misclassification between PCV and wetAMD.

Concerning the two single-modal models, OCT-CNN outperforms CFP-CNN, 0.886 \textit{vs} 0.774 in terms of the overall F1 score. The confusion matrices in Fig. \ref{fig:cm} reveal more details. CFP-CNN recognizes the normal class as easy as OCT-CNN. It also has a good ability to find true PCV. However, CFP-CNN tends to incorrectly predict dryAMD as wetAMD and wetAMD as PCV. OCT-CNN also has difficulty in distinguishing PCV from wetAMD. Due to its relatively distinct visual patterns, dryAMD is relatively easier to be recognized than PCV and wetAMD, thus better separated in the learned CNN feature spaces, see Fig. \ref{fig:tsne}.  Hence, although dryAMD has fewer training samples, its result is better than that of the other two.

\input{fig-tsne}

The better overall performance of OCT-CNN-\textit{da} against OCT-CNN (and \revise{CFP}-CNN-\textit{da} against \revise{CFP}-CNN) suggests that the single-modal models also  benefit from the synthetic instances.
%
%\wsrevise{Note that the proposed data augmentation strategies achieve more improvement on MM-CNN, because it’s a complex model with more parameters to be trained, which is likely to be overfit without strong data augmentation.}
As Table 3 shows, \revise{CFP}-CNN-da has higher sensitivity for dryAMD (0.817) and PCV (0.667) than \revise{CFP}-CNN (0.783 and 0.533). This, however, comes at the cost of losing sensitivity in wetAMD (from 0.550 to 0.417). More examples of wetAMD are misclassified. \revise{Recall that for both human and CNNs, the CFP modality alone is inadequate for recognizing wetAMD. We tried to train CFP-CNN on the synthetic data without finetuning on real data. That model recognizes wetAMD at sensitivity of 0.417, specificity of 0.928 and F1 of 0.547. The performance, though lower than that of CFP-CNN trained on real data, remains meaningful. We therefore attribute the performance degeneration of wetAMD to the unreliability of the CFP modality other than the CAM-conditioned image synthesis}. The data augmentation is in general beneficial for \revise{CFP-CNN and OCT-CNN}.
%, improving its overall F1 from 0.774 to 786 and Accuracy from 0.717 to 0.725.

\revise{Compared to the single-modal CNNs, the amount of trainable parameters in MM-CNN is doubled, from 11.2M to 22.4M. This makes MM-CNN more data-hungry and more likely to  overfit when learning from the small amount of real multi-modal samples, thus worse than OCT-CNN. For this reason, the proposed data augmentation brings more improvement to MM-CNN than the single-modal baselines.} 
\revise{MM-MHSA, despite its excellent performance in various CV tasks \cite{yu2019multimodal}, is not on par with MM-CNN for the current task, where multi-modal training samples are in short supply.} 
%Predictions of some test data 
Some qualitative results are shown in Fig. \ref{fig:instance}.

%Both quantitative and qualitative results justify the effectiveness of the proposed solution.

%for multi-modal AMD categorization.

%% file: table_result_s_vs_m.tex
%\caption{\textbf{Comparing single-modal and multi-modal models}.  We use OCT-CNN, the best single-modal baseline, as a reference. Relative improvements over this reference are shown in parentheses. Our proposed MM-CNN-\textit{da} performs the best.}
%\label{tab:s_v_m}

%\caption{\textbf{Zoom-in view of Table \ref{tab:s_v_m}}, in terms of Sensitivity (\textit{Sen.}) and Specificity (\textit{Spe.}).}
%\label{tab:s_v_m_detail}

\begin{table*}[tbh!]
\normalsize
\renewcommand\arraystretch{1.1}
\centering
\caption{\textbf{Comparing single-modal and multi-modal models}. We use OCT-CNN, the best single-modal baseline, as a reference. Relative improvements over OCT-CNN are shown in parentheses. \textit{Sen.} indicates Sensitivity and \textit{Spe.} as Specificity. For each model we run training and evaluation three times and report average scores per metric. Our proposed MM-CNN-\textit{da} performs the best. }
\label{tab:s_v_m}
\scalebox{0.73}{
%\begin{tabular}{@{}|l | r | r | r | r | l|l |@{}}
\begin{tabular}{@{}|l | r|r|r | r|r|r | r|r|r | r|r|r | l|l|@{}}
\hline

&\multicolumn{3}{c|}{\textbf{normal}}  & \multicolumn{3}{c|}{\textbf{dryAMD}}  & \multicolumn{3}{c|}{\textbf{PCV}} & 			\multicolumn{3}{c|}{\textbf{wetAMD}} & \multicolumn{2}{c|}{\textbf{Overall}}\\
%\cmidrule{3-5} \cmidrule(l){6-8} \cmidrule(l){9-11} \cmidrule(l){12-14} \cmidrule(l){15-16}
\cline{2-15}
\textbf{Model} & \textit{Sen.}  & \textit{Spe.}  & \textit{F1.} & \textit{Sen.}  & \textit{Spe.} & \textit{F1.} & \textit{Sen.}  & \textit{Spe.} & \textit{F1.}  & \textit{Sen.}  & \textit{Spe.} & \textit{F1.} & \textit{F1.}  & \textit{Accuracy}\\

\hline
%& \textbf{1.000} & \textbf{1.000}  & 0.783 & 0.867  & 0.533 & 0.928  & 0.550 & 0.828 
\revise{CFP}-CNN 
& \textbf{1.000} & \textbf{1.000} & \textbf{1.000} 
& 0.783 & 0.867 & 0.798 
& 0.533 & 0.928 & 0.636 
& 0.550 & 0.828  & 0.661  
& 0.774 ($\downarrow$\textcolor{darkgreen}{-12.6\%}) & 0.717 ($\downarrow$\textcolor{darkgreen}{-12.3\%})  \\
%& \textbf{1.000} & 0.989  & 0.817 & 0.867  & 0.667 & 0.906  & 0.417 & 0.872 \\
\revise{CFP}-CNN-\textit{da} 
&\textbf{1.000} & 0.989 & 0.994 
& 0.817 & 0.867  & 0.831 
& 0.667 & 0.906  & 0.764  
& 0.417 & 0.872 & 0.555 
& 0.786 ($\downarrow$\textcolor{darkgreen}{-11.3\%}) & 0.725 ($\downarrow$\textcolor{darkgreen}{-11.4\%})  \\
%& \textbf{1.000} & \textbf{1.000}  & 0.877 & 1.000  & 0.745 & 0.889  & 0.754 & 0.854 \\
OCT-CNN 
& \textbf{1.000} & \textbf{1.000} & \textbf{1.000}  
& 0.877 & 1.000 & 0.934  
& 0.745 & 0.889 & 0.810  
& 0.754 & 0.854 & 0.801 
& 0.886 
& 0.818  \\
%& \textbf{1.000} & 0.992  & 0.868 & 0.994  & 0.766 & 0.910  & 0.763 & 0.860 \\
OCT-CNN-\textit{da}  
& \textbf{1.000} & 0.992 & 0.996
& 0.868 & 0.994 & 0.927  
& 0.766 & 0.910 & 0.831  
& 0.763 & 0.860 & 0.809 
& 0.891 ($\uparrow$\textcolor{red}{+0.6\%}) & 0.825 ($\uparrow$\textcolor{red}{+0.9\%})  \\

\hline
\revise{EarlyFusion}
& \textbf{1.000} 	 & 0.992 	 & 0.996
& 0.825 	 & 0.997 	 & 0.902
& 0.759 	 & 0.847 	 & 0.796
& 0.640 	 & 0.851 	 & 0.730
& 0.856 ($\downarrow$\textcolor{darkgreen}{-3.4\%}) 
& 0.779 ($\downarrow$\textcolor{darkgreen}{-4.8\%}) \\
LateFusion   
& \textbf{1.000} & \textbf{1.000} & \textbf{1.000} 
& \textbf{1.000} & 0.943 & \textbf{0.971} 
& 0.553 & \textbf{0.948} & 0.698 
& 0.790 & 0.827 & 0.808 
& 0.869 ($\downarrow$\textcolor{darkgreen}{-1.9\%}) 
& 0.792 ($\downarrow$\textcolor{darkgreen}{-3.2\%}) \\
\revise{MM-MHSA}   
& \textbf{1.000} & \textbf{1.000} & \textbf{1.000} 
& 0.763 	 & 0.997 	 & 0.863 
& 0.667 	 & 0.924 	 & 0.753 
& 0.816 	 & 0.771 	 & 0.785 
& 0.850 ($\downarrow$\textcolor{darkgreen}{-4.1\%}) 
& 0.779 ($\downarrow$\textcolor{darkgreen}{-4.8\%}) \\
MM-CNN   	
& \textbf{1.000} & \textbf{1.000} & \textbf{1.000}  
& 0.737 & 0.997 & 0.847  
& 0.787 & 0.910 & 0.841  
& 0.790 & 0.819 & 0.801 
& 0.872 ($\downarrow$\textcolor{darkgreen}{-1.6\%}) & 0.804 ($\downarrow$\textcolor{darkgreen}{-1.7\%})  \\
%& \textbf{1.000} & \textbf{1.000}  & 0.868 & \textbf{1.000}  & \textbf{0.794} & \textbf{0.948}  & \textbf{0.868} & \textbf{0.860}
MM-CNN-\textit{da} 
& \textbf{1.000} & \textbf{1.000} & \textbf{1.000} 
& 0.868 & \textbf{1.000} & 0.929  
& \textbf{0.794} & \textbf{0.948} & \textbf{0.864}  
& \textbf{0.868} & \textbf{0.860} & \textbf{0.864} 
& \textbf{0.914} ($\uparrow$\textcolor{red}{+3.2\%})	& \textbf{0.863} ($\uparrow$\textcolor{red}{+5.5\%}) \\
\hline
\end{tabular}
}
\end{table*}

%% file: fig-tsne.tex
\begin{figure}[tbh!]
  \centering
  \subfigure[CFP-CNN feature space]{
  \includegraphics[width=0.47\columnwidth]{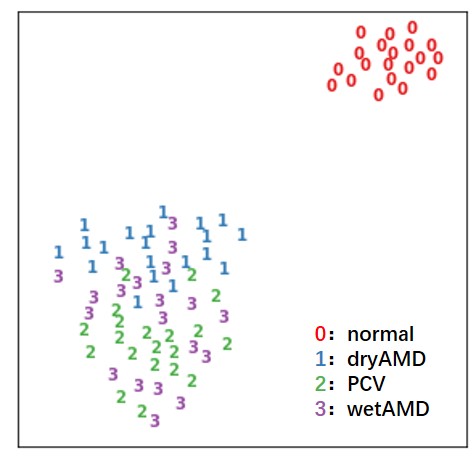}}
  \subfigure[OCT-CNN feature space]{
  \includegraphics[width=0.47\columnwidth]{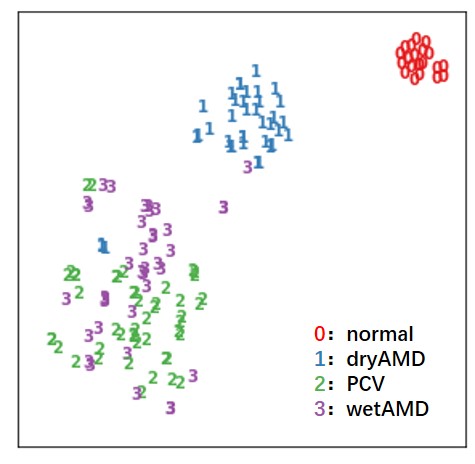}}
  \caption{\revise{\textbf{Visualization of test image distribution in specific CNN feature spaces by t-SNE  \cite{van2008visualizing}}.}}
  \label{fig:tsne}
  \end{figure}

%% file: eval-da.tex
In the previous experiment we report the performance obtained by the joint use of the two data augmentation methods. We now present an ablation study to reveal the effect of the individual method and the training strategy. We also investigate the advantage of the proposed CAM-based image synthesis against the state-of-the-art alternative~\cite{burlina2019assessment}.

\textbf{Effect of the individual data augmentation methods}. Table \ref{tab:da-res} shows the effect of different data augmentation methods and their combinations on the performance of multi-modal CNNs. Note that common data augmentation is \revise{applied to all methods}. For instance, the performance of Loose Pairing in the third row is actually obtained by a joint use of common data augmentation and Loose Pairing, while ``Loose Pairing + CAM-conditioned image synthesis'' is a combination of common data augmentation, Loose Pairing and CAM-conditioned image synthesis. As shown in Table \ref{tab:da-res}, using either Loose Pairing or CAM-conditioned image synthesis is beneficial, bringing in relative improvements of 2.9\% and 2.3\% in F1 and 4.1\% and 5.2\% in Accuracy, respectively, when compared to MM-CNN that uses only common data augmentation. The best combination is to use CAM-conditioned image synthesis to expand the single-modal samples, Loose Paring to expand the multi-modal samples, followed by the two-stage training strategy to properly balance the synthetic and real samples. Such a strategy gives relative improvements of 4.8\% in F1 and 7.3\% in Accuracy.

\begin{table}[tb!]
\normalsize
\renewcommand\arraystretch{1}
\centering
\caption{\textbf{Performance of MM-CNN trained with various data augmentation methods}, sorted in ascending order in terms of F1. We use MM-CNN, exclusively trained with common data augmentation, as a reference. Relative improvements over MM-CNN are shown in parentheses. While all data augmentation methods are helpful, the joint use of loose pairing and CAM-conditioned image synthesis is the best.}
\label{tab:da-res}
\scalebox{0.73}{
\begin{tabular}{@{}|l|l|l|@{}}
\hline
\textbf{Data augmentation} & \textbf{F1}  & \textbf{Accuracy}\\
\hline
%\specialcell{\wsrevise{$\circ$ Common data augmentation}\\ 
%                    \wsrevise{$\circ$ CAM-conditioned image synthesis} \\
%                    (\wsrevise{Use only abnormal synthetic images} \\ \wsrevise{and normal real images})}                 
%                    & 0.826 ($\downarrow$\textcolor{darkgreen}{-5.3\%})
%                    & 0.741 ($\downarrow$\textcolor{darkgreen}{-7.8\%})\\
%\hline
\specialcell{$\circ$ Common data augmentation \\(MM-CNN in Table \ref{tab:s_v_m})} & 0.872  & 0.804  \\
\hline
\specialcell{\revise{$\circ$ Common data augmentation}\\ $\circ$ Label-conditioned image synthesis}             & 0.887 ($\uparrow$\textcolor{red}{+1.7\%}) & 0.825 ($\uparrow$\textcolor{red}{+2.6\%})\\
\hline
\specialcell{\revise{$\circ$ Common data augmentation}\\ $\circ$ CAM-conditioned image synthesis}                 & 0.892 ($\uparrow$\textcolor{red}{+2.3\%})& 0.846 ($\uparrow$\textcolor{red}{+5.2\%})\\
\hline
\specialcell{\revise{$\circ$ Common data augmentation}\\ $\circ$ Loose Pairing \\(Our conference version~\cite{wang2019two})} & 0.897 ($\uparrow$\textcolor{red}{+2.9\%}) & 0.837 ($\uparrow$\textcolor{red}{+4.1\%})\\
\hline
\specialcell{\revise{$\circ$ Common data augmentation}\\ $\circ$ Loose Pairing + \\CAM-conditioned image synthesis,\\ \textit{one-stage training}}  & 0.904 ($\uparrow$\textcolor{red}{+3.7\%})& 0.849 ($\uparrow$\textcolor{red}{+5.6\%})\\
\hline
\specialcell{\revise{$\circ$ Common data augmentation}\\ $\circ$ Loose Pairing + \\CAM-conditioned image synthesis,\\ \textit{two-stage training}}  & \textbf{0.914} ($\uparrow$\textcolor{red}{+4.8\%})& \textbf{0.863} ($\uparrow$\textcolor{red}{+7.3\%})\\
\hline
\end{tabular}
}
\end{table}

%, both Loose Pairing and CAM-conditioned image synthesis are effective when used alone. Compared to MM-CNN, which is trained using common data augmentation only, the two methods bring in relative improvements of 2.9\% and 2.3\% in F1 and 4.1\% and 5.2\% in Accuracy, respectively.

\textbf{Training strategy}. To justify the necessity of the two-stage training strategy described in Section \ref{ssec:training}, we instead train the two-stream CNN on the real and synthetic data simultaneously, with a ratio of 1:3. The performance of this one-stage training strategy is provided in the \revise{second last} row of Table \ref{tab:da-res}. While also improving over MM-CNN, it is not on a par with the two-stage training strategy.

\begin{figure*}[tb!]
\centering
\subfigure[Label-conditioned GAN]{
\includegraphics[width=1.8\columnwidth]{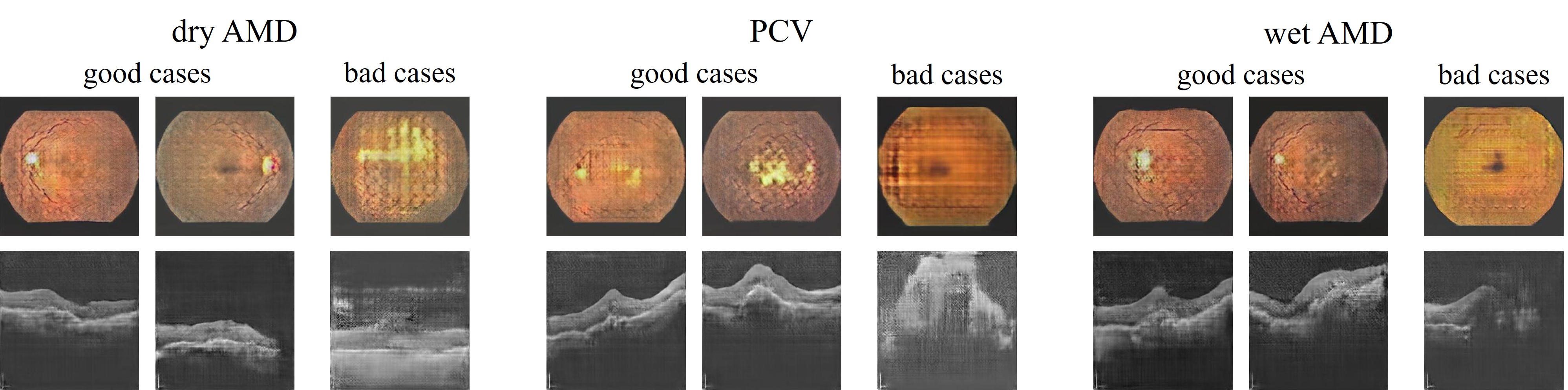}}
\subfigure[CAM-conditioned GAN]{
\includegraphics[width=1.8\columnwidth]{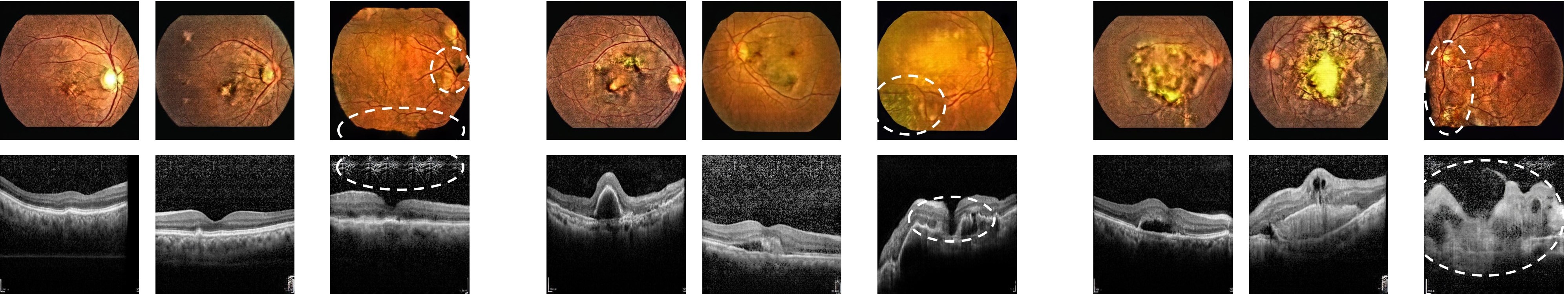}}
%\subfigure[Some Low-quality Examples of CAM-conditioned GAN]{
%\label{fig:gancmp:c}
%\includegraphics[width=2\columnwidth]{images/bad_syn}}
\caption{\textbf{Images generated by label-conditioned / CAM-conditioned GANs}. The latter generates images that are visually more realistic and show better details around the macula. For our method, a synthetic image can be paired with its source image (from which CAMs are extracted), whilst a synthetic image by label-conditioned GAN can only be paired with a class label and random noise. So a controlled comparison given the same source image is unfeasible. Failed cases are shown in the last column in each group, with unrealistic regions marked out by dotted ellipses.}
\label{fig:gancmp}
\end{figure*}

\textbf{Alternative for image synthesis}. As we have noted in Section \ref{ssec:related-da}, Burlina \etal~\cite{burlina2019assessment} use PGGAN to generate \revise{CFP} images for binary AMD categorization. In order to use PGGAN to generate \wsrevise{CFP} / OCT images for each of the three AMD subclasses in this study, we construct the input of PGGAN by concatenating a 3-dimensional one-hot label vector and a random noise vector. Following~\cite{alec2016unsupervised}, we set the size of the input vector to be 100. We term this baseline label-conditioned image synthesis. As Table \ref{tab:da-res} shows, the baseline is less effective than CAM-conditioned image synthesis. Note that our proposed method is capable of explicitly leveraging the spatial information of abnormal regions in a source image in the form of CAMs (Fig. \ref{fig:syn}). By contrast, the input of the baseline is limited to the class label and the random noise. Some synthetic images are presented in Fig. \ref{fig:gancmp}. Compared to images generated by the label-conditioned GAN, images generated by our CAM-conditioned GAN are visually more realistic with better details, in particular around the macular. We also observe failed cases that contain unrealistic regions, see the last column of each group in Fig. \ref{fig:gancmp}. Nonetheless, such base cases are not common. As the confusion matrices in Fig. \ref{fig:cm} show, the synthetic training examples contribute to AMD categorization by reducing misclassification between PCV and wetAMD. \revise{To assess the overall quality of the synthetic data, we train a four-class MM-CNN  on the synthetic data plus real images of the normal class. With F1 of 0.826 and Accuracy of 0.741, the model is somewhere between CFP-CNN and OCT-CNN, suggesting that the synthetic data is largely meaningful.}

%\ws{
%  \textbf{The best combination of strategies}. The state-of-the-art can be obtained by combining Loose Pairing, CAM-conditional synthesis, the two-stage training strategy, and the use of CLAHE. Specifically, Loose Pairing and CAM-conditional synthesis expand the training set, two-stage training alleviates the problem of imbalance between real and synthetic images, and CLAHE contributes to high-quality synthesis of color fundus. Comparing with using common data augmentation only, the strategies combination (also using common data augmentation) obtains relative improvements of 4.8\% in F1 and 7.3\% in Accuracy.
%}

%% file: eval-sota.tex
\textbf{Baselines}. 
%As aforementioned, the only existing work on multi-modal AMD categorization is by Yoo \etal \cite{YooThe}, where the authors employ VGG19 pretrained on ImageNet to extract visual features from fundus and OCT images and then train a random forest classifier on strictly matched pairs. As their data is not publicly available, we replicate their method and evaluate on our test set. For a fair comparison, we substitute ResNet-18 for VGG19. Moreover, we investigate if the proposed data augmentation is also beneficial for the baseline. 
As we have noted in Section \ref{sec:intro}, the only prior art on multi-modal AMD categorization is by Yoo \etal \cite{YooThe}. As their experimental data is non-public, we replicate their method, \ie feature extraction by a pretrained CNN and classification by a random forest classifier. To make the comparison fair, we replace VGG19 used in ~\cite{YooThe} by ResNet-18. In addition, we are interested in whether the proposed multi-modal data augmentation method is also useful for this baseline. We term this enhanced baseline Yoo \etal -\textit{da}.

\textbf{Results}. 
As Table \ref{tab:sota} shows, the proposed MM-CNN-\textit{da} clearly outperforms Yoo \etal, 0.914 \textit{vs} 0.792 in F1 and 0.863 \textit{vs} 0.690 in Accuracy. the proposed data augmentation methods also work for Yoo \etal, brining in relative improvements of 5.7\% in F1 and 7.8\% in Accuracy, respectively. Still, Yoo \etal-\textit{da} is less effective than MM-CNN-\textit{da}. The result justifies the superiority of our end-to-end learning method. 
%for multi-modal AMD categorization. 

\begin{table}[tbh!]
\normalsize
\renewcommand\arraystretch{1.1}
\centering
\caption{\textbf{Comparison with SOTA for multi-modal AMD categorization}.}
\label{tab:sota}
\scalebox{0.73}{
\begin{tabular}{@{}|l | r | r | r | r | r|r |@{}}
\hline
\multirow{2}{*}{\textbf{Model}}    & \multicolumn{5}{c|}{\textbf{F1-score}}  & \multirow{2}{*}{\textbf{Accuracy}} \\
	\cline{2-6}
  & \textit{normal} & \textit{dryAMD}  & \textit{PCV}  & \textit{wetAMD} & \textit{Overall} & \\

\hline
Yoo \etal \cite{YooThe} & \textbf{1.000} & 0.783 & 0.648 & 0.736 & 0.792 & 0.690 \\
\hline
Yoo \etal-\textit{da} 	& \textbf{1.000} & 0.880 & 0.740 & 0.726 & 0.837 & 0.744 \\
\hline 
MM-CNN-\textit{da} 		& \textbf{1.000} & \textbf{0.929} & \textbf{0.864} & \textbf{0.864} & \textbf{0.914}  & \textbf{0.863} \\
\hline
\end{tabular}
}
\end{table}

%% file: discussion.tex
There are several limitations in the current study. Due to the tremendous cost of collecting multi-modal data with ground-truth, the dataset used in our experiments is relatively small. While this does not affect our major conclusions (see the Appendix), the performance of our model is likely to drop when directly applying it on external data collected from different devices. Although we believe that our multi-modal approach can go beyond AMD categorization, its applicability for recognizing more retinal diseases remains to be justified. We have release our source code\footnoteref{github} so readers of interest may train and evaluate based on their own data.

Despite its effectiveness for improving classification, CAM-based image synthesis is not guaranteed to generate clinically and physiologically valid images. For instance, the fundus image generated with manipulated CAM on the first row of Fig. \ref{fig:syn-cfp} appears to have two macular structures. A hotfix to deal with the issue is probably to detect primary physiological structures such as macula and optic disc \cite{mlmi2019-uwf} on the generated images and heuristically exclude the invalid.

Our multi-modal framework accepts only one single OCT image per input. So there is a gap between our setup and the real clinical settings, where an OCT scan typically renders a 3D volume, and very often the operator will choose more than one B-scan from the volume. To handle the OCT volume as input, a simple way is to select the central B-scan, with the other B-scans ignored. Alternatively, we may consider pairing each of the B-scans with the CFP image, using the current framework to make predictions per pair, and finally 
adopting certain score aggregation strategy, say max pooling, to make the volume-level prediction. We leave this for future work.

%% file: append.tex
\textbf{The effect of CLAHE}. Table \ref{tab:clahe} shows the performance of models with and without the CLAHE enhancement on color fundus images. Concerning the overall performance, CLAHE is helpful for CFP-CNN and MM-CNN-\textit{da}, while no positive nor negative effect is observed on MM-CNN.

\begin{table}[tbh!]
    \normalsize
    \renewcommand\arraystretch{1.1}
    \centering
    \caption{\textbf{Performance of models with (+) / without (--) CLAHE}. }

    \label{tab:clahe}
    \scalebox{0.57}{
    \begin{tabular}{@{}|l | c | r | r | r | r | l|l |@{}}
    \hline
    \multirow{2}{*}{\textbf{Model}}  & \multirow{2}{*}{\textbf{CLAHE}}  & \multicolumn{5}{c|}{\textbf{F1-score}}  & \multirow{2}{*}{\textbf{Accuracy}} \\
      \cline{3-7}
     & & \textit{normal} & \textit{dryAMD}  & \textit{PCV}  & \textit{wetAMD} & \textit{Overall} & \\
    
    \hline
    \multirow{2}{*}{\revise{CFP-CNN}} & -- & 1.000 & 0.816 & 0.663 & 0.570 & 0.762 & 0.700 \\
    \cline{2-8}
    & + & 1.000 & 0.798 & 0.636 & 0.661 & 0.774 ($\uparrow$\textcolor{red}{+1.6\%})& 0.717 ($\uparrow$\textcolor{red}{+2.4\%})\\
    \hline
    \multirow{2}{*}{MM-CNN} & --	& 1.000 & 0.859 & 0.818 & 0.809 & 0.872 & 0.804 \\
    \cline{2-8}
    & + & 1.000 & 0.847 & 0.841 & 0.801 & 0.872 ($\quad$\textcolor{gray}{0.0\%})& 0.804 ($\quad$\textcolor{gray}{0.0\%})\\
    \hline 
    \multirow{2}{*}{MM-CNN-\textit{da}}	&	-- & 0.989 & 0.882 & 0.858 & 0.860 & 0.839 & 0.897 \\
    \cline{2-8}
    & + & 1.000 & 0.929 & 0.864 & 0.864 & 0.863 ($\uparrow$\textcolor{red}{+2.9\%})& 0.914 ($\uparrow$\textcolor{red}{+2.5\%})\\
    \hline
    \end{tabular}
    }
    \end{table}

\begin{figure}[tbh!]
  \centering
  \includegraphics[width=0.97\columnwidth]{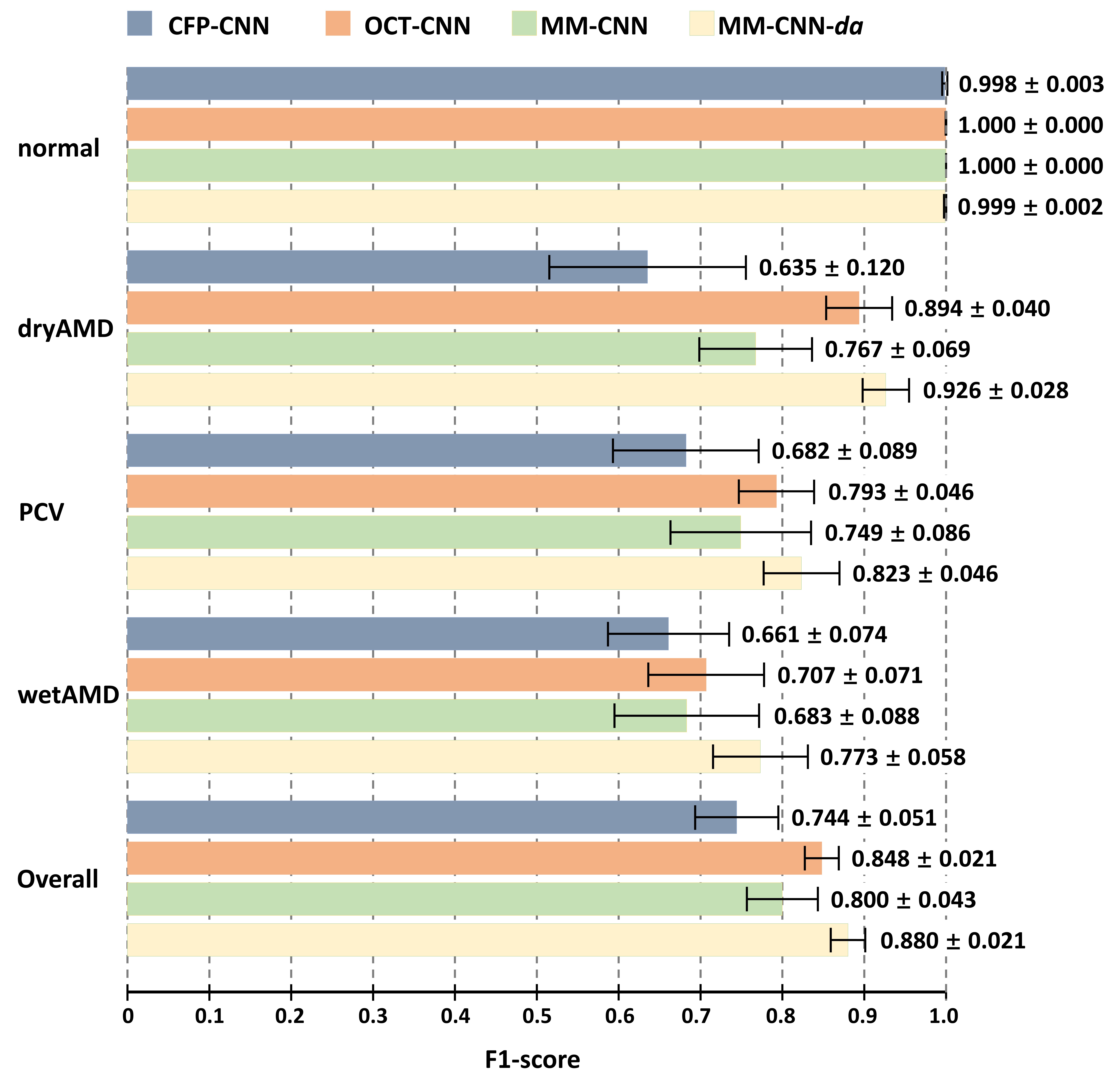}
  \caption{\textbf{Model performance given five distinct data splits}.
  }
  \label{fig:bar}
  \end{figure}

 \textbf{The effect of distinct data splits}. So far all the experiments are performed based on the data split provided in Table \ref{tab:data}. To check whether our major conclusions depend on this specific data split, we re-run our experiments on four new data splits, each obtained by shuffling the entire dataset followed by training / validation / test partition as described in  Sec. \ref{ssec:expset}. Viewing the data split as a random factor, averaging the performance of a model obtained on distinct data splits allows us to cancel out this random factor. As shown in Fig. \ref{fig:bar}, our major conclusions, \ie OCT-CNN is better than \revise{CFP}-CNN, MM-CNN alone is insufficient, and MM-CNN-\textit{da} is the best, remain valid. In addition, we observe that the relatively small amount of dryAMD for training mainly affects \revise{CFP}-CNN, which has a relatively large standard deviation (std) of 0.120 in F1 for dryAMD. In contrast, OCT-CNN has a much smaller std of 0.040, largely because dryAMD-related visual patterns such as drusen are more visible in OCT images.

We have also tried with subject-level split,  adjusting the previous data split in Table \ref{tab:data} to ensure zero subject overlap between the training / validation / test sets, see Table \ref{tab:data_splitP}. The performance of the primary models on the new data split is presented in Table \ref{tab:re_splitP}. Our major conclusions, \ie the two-stream network architecture alone is insufficient to beat the best single-modal baseline and the proposed multi-modal data augmentation is effective, remain valid.

 \input{table_data_splitP}

 \input{table_results_splitP}

%% file: table_data_splitP.tex
\begin{table}[tbh!]
    \renewcommand{\arraystretch}{1}
    \centering
    \caption{\revise{\textbf{Subject-level data split}.  In parentheses are number of subjects per class.} }\label{tab:data_splitP}
    \scalebox{0.8}{
    \begin{tabular}{|l | r|r | r|r | r|r |}
    \hline
       
    \multirow{2}{*}{\textbf{Class}} & \multicolumn{2}{c|}{\textbf{Training set}} & \multicolumn{2}{c|}{\textbf{Validation set}}  & \multicolumn{2}{c|}{\textbf{Test set}}  \\
     \cline{2-7}
    
    & \textit{\wsrevise{CFP}} 	& \textit{OCT} 	& \textit{\wsrevise{CFP}} 	& \textit{OCT}	& \textit{\wsrevise{CFP}} 	&  \textit{OCT} \\
    \hline
    \textit{normal} 	& 155 (155)	& 156 (155)	& 20 (20)	& 20 (20)	& 20 (20)	& 20 (20) \\
    \hline
    \textit{dryAMD} 	& 67 (57)	& 42 (23)	& 20 (19)	& 36 (19)	& 20 (18)	& 28 (17) \\
    \hline
    \textit{PCV} 	& 259 (228)	& 294 (137)	& 20 (20)	& 43 (20)	& 20 (19)	& 43 (19) \\
    \hline
    \textit{wetAMD} 	& 453 (340)	& 532 (245)	& 20 (20)	& 35 (19)	& 20 (19)	& 40 (19) \\
    \hline
    \end{tabular}
    }
    \end{table}

%% file: table_results_splitP.tex
\begin{table}[tbh!]
\normalsize
\renewcommand\arraystretch{1}
\centering
\caption{\revise{\textbf{Performance of single-modal / multi-modal models based on the data split in Table \ref{tab:data_splitP}.}}}
\label{tab:re_splitP}
\scalebox{0.66}{
\begin{tabular}{@{}|l | r | r | r | r | r|r |@{}}
\hline
\multirow{2}{*}{\textbf{Model}}    & \multicolumn{5}{c|}{\textbf{F1-score}}  & \multirow{2}{*}{\textbf{Accuracy}} \\
	\cline{2-6}
  & \textit{normal} & \textit{dryAMD}  & \textit{PCV}  & \textit{wetAMD} & \textit{Overall} & \\

\hline
CFP-CNN & 0.997	& 0.732	& 0.824	& 0.643	& 0.799	& 0.733 \\
\hline
OCT-CNN & \textbf{1.000}	& 0.849	& 0.869	& 0.789	& 0.877	& 0.819 \\
\hline 
LateFusion	& \textbf{1.000}	& 0.813	& 0.919	& 0.74	& 0.868	& 0.807 \\
\hline 
MM-CNN	& \textbf{1.000}	& 0.800	& 0.852	& 0.680	& 0.833	& 0.766 \\
\hline
MM-CNN-\textit{da} & \textbf{1.000}	& \textbf{0.904}	& \textbf{0.913}	& \textbf{0.823}	& \textbf{0.910}	& \textbf{0.863} \\
\hline
\end{tabular}
}
\end{table}

%% file: main.bbl
\begin{thebibliography}{10}
\providecommand{\url}[1]{#1}
\csname url@samestyle\endcsname
\providecommand{\newblock}{\relax}
\providecommand{\bibinfo}[2]{#2}
\providecommand{\BIBentrySTDinterwordspacing}{\spaceskip=0pt\relax}
\providecommand{\BIBentryALTinterwordstretchfactor}{4}
\providecommand{\BIBentryALTinterwordspacing}{\spaceskip=\fontdimen2\font plus
\BIBentryALTinterwordstretchfactor\fontdimen3\font minus
  \fontdimen4\font\relax}
\providecommand{\BIBforeignlanguage}[2]{{%
\expandafter\ifx\csname l@#1\endcsname\relax
\typeout{** WARNING: IEEEtran.bst: No hyphenation pattern has been}%
\typeout{** loaded for the language `#1'. Using the pattern for}%
\typeout{** the default language instead.}%
\else
\language=\csname l@#1\endcsname
\fi
#2}}
\providecommand{\BIBdecl}{\relax}
\BIBdecl

\bibitem{Wan2014Global}
W.~L. Wong \textit{et~al.}, ``Global prevalence of age-related macular
  degeneration and disease burden projection for 2020 and 2040: A systematic
  review and meta-analysis,'' \textit{The Lancet Global Health}, vol.~2, no.~2,
  pp. e106--e116, 2014.

\bibitem{dewan2006htra1}
A.~DeWan \textit{et~al.}, ``{HTRA1} promoter polymorphism in wet age-related
  macular degeneration,'' \textit{Science}, vol. 314, no. 5801, pp. 989--992,
  2006.

\bibitem{lee2017efficacy}
J.~E. Lee \textit{et~al.}, ``Efficacy of fixed-dosing aflibercept for treating
  polypoidal choroidal vasculopathy: 1-year results of the vault study,''
  \textit{Graefe's Archive for Clinical and Experimental Ophthalmology}, vol.
  255, no.~3, pp. 493--502, 2017.

\bibitem{laude2010polypoidal}
A.~Laude \textit{et~al.}, ``Polypoidal choroidal vasculopathy and neovascular
  age-related macular degeneration: Same or different disease?''
  \textit{Progress in Retinal and Eye Research}, vol.~29, no.~1, pp. 19--29,
  2010.

\bibitem{wong2016age-related}
C.~W. Wong \textit{et~al.}, ``Age-related macular degeneration and polypoidal
  choroidal vasculopathy in {Asians},'' \textit{Progress in Retinal and Eye
  Research}, vol.~53, pp. 107--139, 2016.

\bibitem{kokame2019anti}
G.~T. Kokame, T.~E. deCarlo, K.~N. Kaneko, J.~N. Omizo, and R.~Lian,
  ``Anti--vascular endothelial growth factor resistance in exudative macular
  degeneration and polypoidal choroidal vasculopathy,'' \textit{Ophthalmology
  Retina}, vol.~3, no.~9, pp. 744--752, 2019.

\bibitem{de2018clinically}
J.~De~Fauw \textit{et~al.}, ``Clinically applicable deep learning for diagnosis
  and referral in retinal disease,'' \textit{Nature medicine}, vol.~24, no.~9,
  pp. 1342--1350, 2018.

\bibitem{Burlina2016Detection}
P.~Burlina, D.~E. Freund, N.~Joshi, Y.~Wolfson, and N.~M. Bressler, ``Detection
  of age-related macular degeneration via deep learning,'' in \textit{ISBI},
  2016.

\bibitem{Burlina2017Automated}
P.~M. Burlina, N.~Joshi, M.~Pekala, K.~D. Pacheco, D.~E. Freund, and N.~M.
  Bressler, ``Automated grading of age-related macular degeneration from color
  fundus images using deep convolutional neural networks,'' \textit{JAMA
  Ophthalmology}, vol. 135, no.~11, pp. 1170--1176, 2017.

\bibitem{Grassmann2018A}
F.~Grassmann \textit{et~al.}, ``A deep learning algorithm for prediction of
  age-related eye disease study severity scale for age-related macular
  degeneration from color fundus photography,'' \textit{Ophthalmology}, vol.
  125, no.~9, pp. 1410--1420, 2018.

\bibitem{Lee2017Deep}
C.~S. Lee, D.~M. Baughman, and A.~Y. Lee, ``Deep learning is effective for
  classifying normal versus age-related macular degeneration {OCT} images,''
  \textit{Ophthalmology Retina}, vol.~1, no.~4, pp. 322--327, 2017.

\bibitem{Karri2017Transfer}
S.~P.~K. Karri, D.~Chakraborty, and J.~Chatterjee, ``Transfer learning based
  classification of optical coherence tomography images with diabetic macular
  edema and dry age-related macular degeneration,'' \textit{Biomedical Optics
  Express}, vol.~8, no.~2, pp. 579--592, 2017.

\bibitem{Treder2018Automated}
M.~Treder, J.~L. Lauermann, and N.~Eter, ``Automated detection of exudative
  age-related macular degeneration in spectral domain optical coherence
  tomography using deep learning,'' \textit{Graefe's Archive for Clinical and
  Experimental Ophthalmology}, vol. 256, no.~2, pp. 259--265, 2018.

\bibitem{YooThe}
T.~K. Yoo, J.~Y. Choi, J.~G. Seo, B.~Ramasubramanian, S.~Selvaperumal, and
  D.~W. Kim, ``The possibility of the combination of {OCT} and fundus images
  for improving the diagnostic accuracy of deep learning for age-related
  macular degeneration: a preliminary experiment,'' \textit{Medical \&
  Biological Engineering \& Computing}, vol.~57, no.~3, pp. 677--687, 2019.

\bibitem{Burlina2017Comparing}
P.~Burlina, K.~D. Pacheco, N.~Joshi, D.~E. Freund, and N.~M. Bressler,
  ``Comparing humans and deep learning performance for grading {AMD}: A study
  in using universal deep features and transfer learning for automated {AMD}
  analysis,'' \textit{Computers in Biology \& Medicine}, vol.~82, pp. 80--86,
  2017.

\bibitem{Russakoff2019Deep}
D.~B. Russakoff, A.~Lamin, J.~D. Oakley, A.~M. Dubis, and S.~Sivaprasad, ``Deep
  learning for prediction of {AMD} progression: A pilot study,''
  \textit{Investigative Ophthalmology \& Visual Science}, vol.~60, no.~2, pp.
  712--722, 2019.

\bibitem{Kermany2018Identifying}
D.~S. Kermany \textit{et~al.}, ``Identifying medical diagnoses and treatable
  diseases by image-based deep learning,'' \textit{Cell}, vol. 172, no.~5, p.
  1122–1131.e9, 2018.

\bibitem{krizhevsky2012imagenet}
A.~Krizhevsky, I.~Sutskever, and G.~E. Hinton, ``{ImageNet} classification with
  deep convolutional neural networks,'' in \textit{NIPS}, 2012.

\bibitem{simonyan2014very}
K.~Simonyan and A.~Zisserman, ``Very deep convolutional networks for
  large-scale image recognition,'' in \textit{ICLR}, 2015.

\bibitem{feichtenhofer2016convolutional}
C.~Feichtenhofer, A.~Pinz, and A.~Zisserman, ``Convolutional two-stream network
  fusion for video action recognition,'' in \textit{CVPR}, 2016.

\bibitem{yu2019multimodal}
J.~Yu, J.~Li, Z.~Yu, and Q.~Huang, ``Multimodal transformer with multi-view
  visual representation for image captioning,'' \textit{IEEE Transactions on
  Circuits and Systems for Video Technology}, vol.~30, no.~12, pp. 4467--4480,
  2019.

\bibitem{Zhou2015Learning}
B.~Zhou, A.~Khosla, A.~Lapedriza, A.~Oliva, and A.~Torralba, ``Learning deep
  features for discriminative localization,'' in \textit{CVPR}, 2016.

\bibitem{wang2018high}
T.-C. Wang, M.-Y. Liu, J.-Y. Zhu, A.~Tao, J.~Kautz, and B.~Catanzaro,
  ``High-resolution image synthesis and semantic manipulation with conditional
  {GANs},'' in \textit{CVPR}, 2018.

\bibitem{wang2019two}
W.~Wang \textit{et~al.}, ``Two-stream {CNN} with loose pair training for
  multi-modal {AMD} categorization,'' in \textit{MICCAI}, 2019.

\bibitem{kanagasingam2014progress}
Y.~Kanagasingam, A.~Bhuiyan, M.~D. Abramoff, R.~T. Smith, L.~Goldschmidt, and
  T.~Y. Wong, ``Progress on retinal image analysis for age related macular
  degeneration,'' \textit{Progress in Retinal and Eye Research}, vol.~38, pp.
  20--42, 2014.

\bibitem{burlina2019assessment}
P.~M. Burlina, N.~Joshi, K.~D. Pacheco, T.~A. Liu, and N.~M. Bressler,
  ``Assessment of deep generative models for high-resolution synthetic retinal
  image generation of age-related macular degeneration,'' \textit{JAMA
  Ophthalmology}, vol. 137, no.~3, pp. 258--264, 2019.

\bibitem{karras2017progressive}
T.~Karras, T.~Aila, S.~Laine, and J.~Lehtinen, ``Progressive growing of {GANs}
  for improved quality, stability, and variation,'' in \textit{ICLR}, 2018.

\bibitem{zheng2018detection}
R.~Zheng \textit{et~al.}, ``Detection of exudates in fundus photographs with
  imbalanced learning using conditional generative adversarial network,''
  \textit{Biomedical Optics Express}, vol.~9, no.~10, pp. 4863--4878, 2018.

\bibitem{shin2018medical}
H.-C. Shin \textit{et~al.}, ``Medical image synthesis for data augmentation and
  anonymization using generative adversarial networks,'' in
  \textit{International Workshop on Simulation and Synthesis in Medical
  Imaging}, 2018.

\bibitem{xing2019adversarial}
Y.~Xing \textit{et~al.}, ``Adversarial pulmonary pathology translation for
  pairwise chest x-ray data augmentation,'' in \textit{MICCAI}, 2019.

\bibitem{zhou2019high}
Y.~Zhou, X.~He, S.~Cui, F.~Zhu, L.~Liu, and L.~Shao, ``High-resolution diabetic
  retinopathy image synthesis manipulated by grading and lesions,'' in
  \textit{MICCAI}, 2019.

\bibitem{yi2019generative}
X.~Yi, E.~Walia, and P.~Babyn, ``Generative adversarial network in medical
  imaging: A review,'' \textit{Medical Image Analysis}, vol.~58, p. 101552,
  2019.

\bibitem{isola2017image}
P.~Isola, J.-Y. Zhu, T.~Zhou, and A.~A. Efros, ``Image-to-image translation
  with conditional adversarial networks,'' in \textit{CVPR}, 2017.

\bibitem{accv2018-laser-scar}
Q.~Wei, X.~Li, H.~Wang, D.~Ding, W.~Yu, and Y.~Chen, ``Laser scar detection in
  fundus images using convolutional neural networks,'' in \textit{ACCV}, 2018.

\bibitem{mmm2019-left-right-eye}
X.~Lai, X.~Li, R.~Qian, D.~Ding, J.~Wu, and J.~Xu, ``Four models for automatic
  recognition of left and right eye in fundus images,'' in \textit{MMM}, 2019.

\bibitem{deng2009imagenet}
J.~Deng, W.~Dong, R.~Socher, L.-J. Li, K.~Li, and L.~Fei-Fei, ``{ImageNet}: A
  large-scale hierarchical image database,'' in \textit{CVPR}, 2009.

\bibitem{Jintasuttisak2014Color}
T.~Jintasuttisak and S.~Intajag, ``Color retinal image enhancement by
  {Rayleigh} contrast-limited adaptive histogram equalization,'' in
  \textit{ICCAS}, 2014.

\bibitem{paszke2019pytorch}
A.~Paszke \textit{et~al.}, ``{PyTorch}: An imperative style, high-performance
  deep learning library,'' in \textit{NeurIPS}, 2019.

\bibitem{van2008visualizing}
L.~Van~der Maaten and G.~Hinton, ``Visualizing data using t-{SNE},''
  \textit{Journal of Machine Learning Research}, vol.~9, no.~11, pp.
  2579--2605, 2008.

\bibitem{alec2016unsupervised}
A.~Radford, L.~Metz, and S.~Chintala, ``Unsupervised representation learning
  with deep convolutional generative adversarial networks,'' in \textit{ICLR},
  2016.

\bibitem{mlmi2019-uwf}
Z.~Yang \textit{et~al.}, ``Joint localization of optic disc and fovea in
  ultra-widefield fundus images,'' in \textit{MLMI@MICCAI}, 2019.

\end{thebibliography}
